\RequirePackage{fix-cm}
\documentclass[smallcondensed,final]{svjour3}[2021]     
\smartqed  
\usepackage{graphicx}
\usepackage{algorithmic}
\usepackage{algorithm}
\usepackage{array}
\usepackage{url}
\usepackage{amsmath}
\usepackage{pdflscape}
\usepackage{times}
\usepackage{graphicx}
\usepackage{latexsym}
\usepackage{amssymb}
\usepackage{xcolor,colortbl}
\usepackage{array}
\usepackage{tikz}
\usepackage{multirow}
\usepackage{booktabs}
\usepackage{color}
\usepackage{tabularx}
\usepackage[draft]{todonotes} 
\usepackage{soul}
\usetikzlibrary{positioning}
\usepackage{fancyhdr}

\newcolumntype{C}[1]{>{\centering\arraybackslash}m{#1}}
\definecolor{Gray}{gray}{0.85}
\definecolor{supergood}{rgb}{.2,.6,.2} 
\definecolor{good}{rgb}{.5,.99,.5} 
\definecolor{equal}{rgb}{.2,.6,.5} 
\definecolor{bad}{rgb}{.9,.7,.1} 
\definecolor{superbad}{rgb}{.9,.1,.1}   
\definecolor{notworking}{rgb}{.5,.5,.5}

\def\makeheadbox{
\scriptsize
This version of the article has been accepted for publication at Artificial Intelligence Review, after peer review but is not the Version of Record and does not reflect post-acceptance improvements, or any corrections. The Version of Record is available online at Springer via \url{https://dx.doi.org/10.1007/s10462-021-10110-3}
}

\begin{document}

\title{A Comparison of Vector Symbolic Architectures}


\author{Kenny Schlegel         \and
        Peer Neubert  \and 
        Peter Protzel
}


\institute{K. Schlegel \at
              Faculty of Electrical Engineering, Chemnitz University of Technology, Germany \\
              Tel.: +49 371 531-32053 \\
              \email{kenny.schlegel@etit.tu-chemnitz.de}           
           \and
           P. Neubert \at
			Faculty of Electrical Engineering, Chemnitz University of Technology, Germany \\
			\email{peer.neubert@etit.tu-chemnitz.de}
			\and           
           P. Protzel \at
           Faculty of Electrical Engineering, Chemnitz University of Technology, Germany \\
           \email{peter.protzel@etit.tu-chemnitz.de}
}

\date{}

\maketitle

\begin{abstract}
Vector Symbolic Architectures combine a high-dimensional vector space with a set of carefully designed operators in order to perform symbolic computations with large numerical vectors. Major goals are the exploitation of their representational power and ability to deal with fuzziness and ambiguity. 
Over the past years, several VSA implementations have been proposed. 
The available implementations differ in the underlying vector space and the particular implementations of the VSA operators.
This paper provides an overview of eleven available VSA implementations and discusses their commonalities and differences in the underlying vector space and operators. 
We create a taxonomy of available binding operations and show an important ramification for non self-inverse binding operations using an example from analogical reasoning. 
A main contribution is the experimental comparison of the available implementations in order to evaluate (1) the capacity of bundles, (2) the approximation quality of non-exact unbinding operations, (3) the influence of combining binding and bundling operations on the query answering performance, and (4) the performance on two example applications: visual place- and language-recognition. 
We expect this comparison and systematization to be relevant for development of VSAs, and to support the selection of an appropriate VSA for a particular task.
The implementations are available. 
\end{abstract}

\section{Introduction}
\label{sec:intro}
This paper is about selecting the appropriate Vector Symbolic Architecture (VSA) to approach a given task. But what is a VSA?
VSAs are a class of approaches to solve computational problems using mathematical operations on large vectors.
A VSA consists of a particular vector space, for example $[-1,1]^D$ with $D=10,000$ (the space of 10,000-dimensional vectors with real numbers between -1 and 1) and a set of well chosen operations on these vectors.
Although each vector from $[-1,1]^D$ is primarily a subsymbolic entity without particular meaning, we can associate a symbolic meaning with this vector. To some initial atomic vectors, we can assign a meaning. For other vectors, the meaning will depend on the applied operations and operands.
This is similar to how a symbol can be encoded in a binary pattern in a computer (e.g., encoding a number). In the computer, imperative algorithmic processing of this binary pattern is used to perform manipulation of the symbol (e.g., do calculations with numbers).
The binary encodings in computers and operations on these bitstrings are optimized for maximum storage efficiency (i.e., to be able to distinguish $2^n$ different numbers in an n-dimensional bitstring) and for exact processing (i.e., there is no uncertainty in the encodings or the outcome of an operation).
Vector Symbolic Architectures follow a considerably different approach:

1) Symbols are encoded in very large atomic vectors, much larger than would be required to just distinguish the symbols. 
VSAs use the additional space to introduce redundancy in the representations, usually combined with distributing information across many dimensions of the vector (e.g., there is no single bit that represents a particular property -- hence a single error on this bit can not alter this property). As an important result, this redundant and distributed representation allows to also store compositional structures of multiple atomic vectors in a vector from the same space.
Moreover, it is known from mathematics that in very high dimensional spaces \textit{randomly} sampled vectors are very likely almost orthogonal~\cite{Kanerva09} (a result of the concentration of measure). This can be exploited in VSAs to encode symbols using \textit{random} vectors and, nevertheless, there will be only a very low chance that two symbols are similar in terms of angular distance measures. 
Very importantly, measuring the angular distance between vectors allows us to evaluate a graded similarity relation between the corresponding symbols.

2) The operations in VSAs are mathematical operations that create, process and preserve the graded similarity of the representations in a systematic and useful way.
For instance, an addition-like operator can overlay vectors and creates a representation that is similar to the overlaid vectors. 
Let us look at an example (borrowed from \cite{Kanerva09}):
Suppose that we want to represent the country \textit{USA} and its properties with symbolic entities -- e.g., the currency \textit{Dollar} and capital \textit{Washington DC} (abbreviated \textit{WDC}).
In a VSA representation, each entity is a high-dimensional vector. For basic entities, for which we do not have additional information to systematically create them, we can use a \textit{random} vector (e.g., sample from $[-1,1]^D$). In our example, these might be \textit{Dollar} and \textit{WDC} -- remember, these two high-dimensional random vectors will be very dissimilar. In contrast, the vector for \textit{USA} shall reflect our knowledge that \textit{USA} is related to \textit{Dollar} and \textit{WDC}.
Using a VSA, a simple approach would be to create the vector for USA as a superposition of the vectors Dollar and WDC by using an operator $+$ that is called bundling: $R_{USA} =  Dollar + WDC$. A VSA implements this operator such that it creates a vector $R_{USA}$ (from the same vector space) that is \textit{similar} to the input vectors -- hence, $R_{USA}$ will be similar to both \textit{WDC} and \textit{Dollar}.

VSAs provide more operators to represent more complex relations between vectors. For instance, a binding operator $\otimes$ that can be used to create role-filler pairs and create and query more expressive terms like: $R_{USA} = Name \otimes USA + Curr \otimes Dollar + Cap \otimes WDC$, with \textit{Name}, \textit{Curr}, and \textit{Cap} being random vectors that encode these three roles. 
Why is this useful? We can now query for the currency of the USA by another mathematical operation (called unbinding) on the vectors and \textit{calculate} the result by: 
$Dollar = R_{USA} \oslash Curr$. Most interestingly, this query would still work under significant amounts of fuzziness -- either due to noise, ambiguities in the word meanings, or synonyms (e.g. querying with \textit{monetary unit} instead of \textit{currency} -- provided that these synonym vectors are created in an appropriate way, i.e. they are similar to some extent).
The following Sec.~\ref{sec:vsa_props} will provide more details on these VSA operators.

Using embeddings in high-dimensional vector spaces to deal with ambiguities is well established in natural language processing \cite{Widdows04}. 
There, the objective is typically a particular similarity structure of the embeddings. 
VSAs make use of a larger set of operations on high-dimensional vectors and focus on the sequence of operations that generated a representation.
A more exhaustive introduction to the properties of these operations can be found in the seminal paper of Pentti Kanerva \cite{Kanerva09} and in the more recent paper \cite{Neubert2019}.
So far, they have been applied in various fields including medical diagnosis \cite{Widdows15}, image feature aggregation \cite{Neubert2021b}, semantic image retrieval \cite{Neubert2021a}, robotics \cite{Neubert2019}, to address catastrophic forgetting in deep neural networks \cite{Cheung19}, fault detection \cite{Kleyko15a}, analogy mapping \cite{Rachkovskij12}, reinforcement learning \cite{Kleyko15}, long-short term memory \cite{Danihelka16}, pattern recognition \cite{Kleyko18}, text classification \cite{Joshi2017}, synthesis of finite state automata \cite{Osipov17}, and for creating hyperdimensional stack machines \cite{Yerxa18}.
Interestingly, also the intermediate and output layers of deep artificial neural networks can provide high-dimensional vector embeddings for symbolic processing with a VSA \cite{Neubert2019,Yilmaz2015,Karunaratne2021}.
Although processing of vectors with thousands of dimensions is currently not very time efficient on standard CPUs, typically, VSA operations can be highly parallelized.
In addition, there are also particularly efficient in-memory implementations of VSA operators possible \cite{Karunaratne2020}.
Further, VSAs support distributed representations, which are exceptionally robust towards noise~\cite{Ahmad15}, an omnipresent problem when dealing with real world data, e.g., in robotics \cite{Thrun05}. 
In the long term, this robustness can also allow to use very power efficient stochastic devices \cite{Rahimi2017} that are prone to bit errors but are very helpful for applications with limited resources (e.g., mobile computing, edge computing, robotics).

As stated initially, a VSA combines a vector space with a set of operations. However, based on the chosen vector space and the implementation of the operations, a different VSA is created. 
In the above list of VSA applications, a broad range of different VSAs has been used. 
They all use a similar set of operations, but the different underlying vector spaces and the different implementations of the operations have a large influence on the properties of each individual VSA.
Basically, each application of a VSA raises the question: Which VSA is the best choice for the task at hand?
This question gained relatively little attention in the literature. 
For instance, \cite{Widdows15}, \cite{Kleyko2018a}, \cite{Rahimi2017} and \cite{Plate1997} describe various possible vector spaces with corresponding bundling and binding operation but do not experimentally compare these VSAs on an application. 
A capacity experiment of different VSAs in combination with Recurrent Neuronal Network memory was done in \cite{Frady2018}. 
However, the authors focus particularly on the application of the recurrent memory rather than the complete set of operators. 

In this paper, we benchmark eleven VSA implementations from the literature. 
We provide an overview of their properties in the following Sec.~\ref{sec:vsa_props}.
This section also presents a novel taxonomy of the different existing binding operators and discusses the algorithmic ramifications of their mathematical properties.
A more practically relevant contribution is the experimental comparison of the available VSAs in Sec.~\ref{sec:exp} with respect to the following important questions:
(1)~\textit{How efficiently can the different VSAs store (bundle) information into one representation?} 
(2)~\textit{What is the approximation quality of non exact unbind operators?} 
(3)~\textit{To what extend are binding and unbinding disturbed by bundled representations?} 
In Sec.~\ref{sec:apps}, we complement this evaluation based on synthetic data with an experimental comparison on two practical applications that involve real-world data: the ability to encode context for visual place recognition on mobile robots and the ability to systematically construct symbolic representations for recognizing the language of a given text.
The paper closes with a summary of the main insights in Sec.~\ref{sec:conclusion}.
Matlab implementations of all VSAs and the experiments are available online.\footnote{\url{https://github.com/TUC-ProAut/VSA_Toolbox}, additional supplementary material is also available at \url{https://www.tu-chemnitz.de/etit/proaut/vsa}}

We want to emphasize the point that a detailed introduction to VSAs and their operators are beyond the scope of this paper -- instead, we focus on a comparison of available implementations. For more basic introductions to the topic please refer to \cite{Kanerva09} or \cite{Neubert2019}.

\section{VSAs and their properties}
\label{sec:vsa_props}

A VSA combines a vector space with a set of operations. The set of operations can vary but typically includes operators for bundling, binding, and unbinding, as well as a similarity measure.
These operators are often complemented by a permutation operator which is important, e.g., to quote information \cite{Gayler98}. Despite their importance, since permutations work very similar for all VSAs, they are not part of this comparison.
Instead we focus on differences between VSAs that can result from differences in one or multiple of the other components described in the following subsections.
We selected the following implementations\footnote{All VSAs are taken from the literature. However, in order to implement and experimentally evaluate them, we had to make additional design decisions for some. This led to the three versions of the MAP architecture from \cite{Gayler98}.} (summarized in Table~\ref{tab:VSAs}): the Multiply-Add-Permute (we use the acronyms \textbf{MAP-C}, \textbf{MAP-B} and \textbf{MAP-I}, to distinguish their three possible variations based on real, bipolar or integer vector spaces) from \cite{Gayler98}, the Binary Spatter Code (\textbf{BSC}) from \cite{Kanerva1996}, the Binary Sparse Distributed Representation from \cite{Rachkovskij2001} (\textbf{BSDC-CDT} and \textbf{BSDC-S} to distinguish the two different proposed binding operations), another Binary Sparse Distributed Representation from \cite{Laiho2015} (\textbf{BSDC-SEG}), the Holographic Reduced Representations (\textbf{HRR}) from \cite{Plate1995} and its realization in the frequency domain (\textbf{FHRR}) from \cite{Plate2003} \cite{Plate94Phd}, the Vector derived Binding (\textbf{VTB}) from \cite{Gosmann2019}, which is also based on the ideas of \cite{Plate94Phd}, and finally an implementation called Matrix Binding of Additive Terms (\textbf{MBAT}) from \cite{Gallant2013}.

All these VSAs share the property of using high-dimensional representations (hypervectors). However, they differ in their specific vector spaces $\mathbb{V}$. 
Section \ref{subsec:hv_elemts} will introduce properties of these high-dimensional vectors spaces and discuss the creation of hypervectors.
The introduction emphasized the importance of a similarity measure to deal with the fuzziness of representations: instead of treating representations as same or different, VSAs typically evaluate their similarity. Sec.~\ref{subsec:sim_meas} will provide details of the used similarity metrics. 
Table~\ref{tab:VSAs} summarizes the properties of the compared VSAs.
In order to solve computational problems or represent knowledge with a VSA, we need a set of operations: bundling will be the topic of Sec.~\ref{subsec:bundling} and binding and unbinding will be explained in Sec.~\ref{subsec:binding}. This section will also introduce a taxonomy that systematizes the significant differences in the available binding implementations.
Finally, Sec.~\ref{subsec:analog} will describe an example application of VSAs to analogical reasoning using the previously described operators.
The application is similar to the USA-representation example from the introduction and will reveal important ramifications of non-self inverse binding operations.

\begin{landscape}
\begin{table}[t]
\caption{Summary of the compared VSAs. $\mathcal{U}(min,max)$ is the uniform distribution in range $[min, max]$. $\mathcal{N}(\mu,\sigma)$ defines the normal distribution with mean $\mu$ and variance $\sigma$. $\mathcal{B}(p)$ represents the Bernoulli distribution with probability $p$. $D$ denotes the number of dimensions and $p$ the density. 
The density $p$ of BSDC architectures is $p\ll1$. \cite{Rachkovskij2001} showed that a probability of $p=\frac{1}{\sqrt{D}}$ results in the largest capacity.
The density of BSDC-SEG corresponds to the number of segments\cite{Laiho2015}. See section~\ref{subsec:hv_elemts} for details.
For each binding and unbinding operator the algebraic properties are listed (associative and commutative) -- either check for true or a cross for false.}
\label{tab:VSAs}
\footnotesize
\begin{center}
\begin{tabular}{p{1.4cm}|p{1.7cm}|p{2.7cm}|C{1.65cm}|C{2cm}|C{1.2cm}|C{1.2cm}|C{1.2cm}|C{1.2cm}|C{0.9cm}}
\hline
\multirow{3}{*}{\textbf{Name}} &  \multirow{3}{*}{\shortstack{\textbf{elements $X$ of} \\ \textbf{vector space $\mathbb{V}$}}} & \multirow{3}{*}{\shortstack{\textbf{Initialization of an} \\ \textbf{atomic vector $x_i$}}} & \multirow{3}{*}{\shortstack{\textbf{typical used} \\ \textbf{Sim. metric}}} & \multirow{3}{*}{\textbf{Bundling}} & \multicolumn{2}{c|}{\textbf{Binding}} & \multicolumn{2}{c|}{\textbf{Unbinding}} & \multirow{3}{*}{\textbf{Ref.}}\\
&&&&& commu-tative & asso-ciative & commu-tative & asso-ciative &  \\

\specialrule{1.5pt}{1pt}{1pt}
\multirow{2}{*}{MAP-C} & \multirow{2}{*}{$X \in\mathbb{R}^D$} & \multirow{2}{*}{$x_i\sim\mathcal{U}(-1,1)$} & \multirow{2}{*}{cosine sim.} &  \multirow{2}{*}{\shortstack{elem. addition \\ with cutting}} & \multicolumn{2}{c|}{elem. multipl.} & \multicolumn{2}{c|}{elem. multipl.} & \multirow{2}{*}{\cite{Gayler98}}\\
&&&&& \checkmark & \checkmark & \checkmark & \checkmark & \\

\hline 
\multirow{2}{*}{MAP-I} & \multirow{2}{*}{$X \in \mathbb{Z}^D$} & \multirow{2}{*}{$x_i\sim\mathcal{B}(0.5) \cdot 2-1$} & \multirow{2}{*}{cosine sim.} &  \multirow{2}{*}{elem. addition} & \multicolumn{2}{c|}{elem. multipl.} & \multicolumn{2}{c|}{elem. multipl.} & \multirow{2}{*}{\cite{Gayler98}}\\
&&&&& \checkmark & \checkmark & \checkmark & \checkmark & \\

\hline
\multirow{2}{*}{HRR} & \multirow{2}{*}{$X \in \mathbb{R}^D$} & \multirow{2}{*}{$x_i\sim\mathcal{N}(0,\frac{1}{D})$} & \multirow{2}{*}{cosine sim.} & \multirow{2}{*}{\shortstack{elem. addition \\ with normalization}} & \multicolumn{2}{c|}{circ. conv.} & \multicolumn{2}{c|}{circ. corr.} &  \multirow{2}{*}{\cite{Plate1995,Plate2003}}\\
&&&&& \checkmark & \checkmark & x & x & \\

\hline
\multirow{2}{*}{VTB} & \multirow{2}{*}{$X \in \mathbb{R}^D$} & \multirow{2}{*}{$x_i\sim\mathcal{N}(0,\frac{1}{D})$} & \multirow{2}{*}{cosine sim.} & \multirow{2}{*}{\shortstack{elem. addition \\ with normalization}} & \multicolumn{2}{c|}{VTB} & \multicolumn{2}{c|}{transpose VTB} &  \cite{Gosmann2019}\\
&&&&& x & x & x & x & \\

\hline
\multirow{2}{*}{MBAT} & \multirow{2}{*}{$X \in \mathbb{R}^D$} & \multirow{2}{*}{$x_i\sim\mathcal{N}(0,\frac{1}{D})$} & \multirow{2}{*}{cosine sim.} & \multirow{2}{*}{\shortstack{elem. addition \\ with normalization}} & \multicolumn{2}{c|}{matrix multipl.} & \multicolumn{2}{c|}{inv. matrix multipl.} &  \cite{Gallant2013}\\
&&&&& x & x & x & x & \\

\hline
\multirow{2}{*}{MAP-B} & \multirow{2}{*}{$X \in\{-1,1\}^D$} & \multirow{2}{*}{$x_i\sim\mathcal{B}(0.5) \cdot 2-1$}  & \multirow{2}{*}{cosine sim.} & \multirow{2}{*}{\shortstack{elem. addition \\ with threshold}} & \multicolumn{2}{c|}{elem. multipl.} & \multicolumn{2}{c|}{elem. multipl.} &  \multirow{2}{*}{\cite{Gayler2009,Kleyko2018}}\\
&&&&& \checkmark & \checkmark & \checkmark & \checkmark & \\

\hline
\multirow{2}{*}{BSC} & \multirow{2}{*}{$X \in \{0,1\}^D$} & \multirow{2}{*}{$x_i\sim\mathcal{B}(0.5)$}  & \multirow{2}{*}{hamming dist.} & \multirow{2}{*}{\shortstack{elem. addition \\ with threshold}} & \multicolumn{2}{c|}{XOR} & \multicolumn{2}{c|}{XOR} &  \multirow{2}{*}{\cite{Kanerva1996}}\\
&&&&& \checkmark & \checkmark & \checkmark & \checkmark & \\

\hline
\multirow{2}{*}{\shortstack{BSDC-CDT}} & \multirow{2}{*}{$X \in \{0,1\}^D$} & \multirow{2}{*}{$x_i\sim\mathcal{B} (p\ll1)$} & \multirow{2}{*}{overlap} & \multirow{2}{*}{disjunction} & \multicolumn{2}{c|}{CDT} & \multicolumn{2}{c|}{-} &  \multirow{2}{*}{\cite{Rachkovskij2001}}\\
&&&&& \checkmark & \checkmark &  &  & \\

\hline
\multirow{2}{*}{BSDC-S} & \multirow{2}{*}{$X \in \{0,1\}^D$} & \multirow{2}{*}{$x_i\sim\mathcal{B}(p\ll1)$} & \multirow{2}{*}{overlap} & \multirow{2}{*}{\shortstack{disjunction \\ (opt. thinning)}} & \multicolumn{2}{c|}{shifting} & \multicolumn{2}{c|}{shifting} &  \multirow{2}{*}{\cite{Rachkovskij2001}}\\
&&&&& x & x & x & x & \\

\hline
\multirow{2}{*}{BSDC-SEG} & \multirow{2}{*}{$X \in \{0,1\}^D$} & \multirow{2}{*}{$x_i\sim\mathcal{B}(p\ll1)$} & \multirow{2}{*}{overlap} &  \multirow{2}{*}{\shortstack{disjunction \\ (opt. thinning)}} & \multicolumn{2}{c|}{segment shifting} & \multicolumn{2}{c|}{segment shifting} &  \multirow{2}{*}{\cite{Laiho2015}}\\
&&&&& \checkmark & \checkmark & x & x & \\

\hline
\multirow{1}{*}{FHRR} & \multirow{1}{*}{$X \in \mathbb{C}^D$} & $x_i=e^{i\cdot \theta}$  & \multirow{1}{*}{angle distance} & \multirow{1}{*}{\shortstack{angles of elem. \\ addition}} & \multicolumn{2}{c|}{elem. angle addition} & \multicolumn{2}{c|}{elem. angle subtraction} & \multirow{2}{*}{\cite{Plate94Phd}}\\
&&$\theta\sim\mathcal{U}(-\pi,\pi)$&&& \checkmark & \checkmark & x & x & \\
\hline
\end{tabular}
\end{center}
\end{table}
\end{landscape}

\subsection{Hypervectors -- The elements of a VSA}
\label{subsec:hv_elemts}

A VSA works in a specific vector space $\mathbb{V}$ with a defined set of operations. 
The generation of hypervectors from the particular vector space $\mathbb{V}$ is an essential step in high-dimensional symbolic processing. 
There are basically three ways to create a vector in a VSA: 
(1) It can be the result of a VSA operation. 
(2) It can be the result of (engineered or learned) encoding of (real-world) data. 
(3) It can be an atomic entity (e.g. a vector that represents a role in a role-filler pair). For these role vectors, it is crucial that they are non-similar to all other unrelated vectors. Luckily, in the high-dimensional vectors spaces underlying VSAs, we can simply use random vectors since they are mutually quasi-orthogonal.
From these three ways, the first will be the topic of the following subsections on the operators.
The second way (encoding other data as vectors, e.g. by feeding an image through a ConvNet)  is part of the section~\ref{subsec:place_recognition} to encode images for visual place recognition. 
The third way of creating basic vectors is topic of this section since it plays an important role when using VSAs and varies significantly for the different available VSAs.

When selecting vectors to represent basic entities (e.g., symbols for which we do not know any relation that we can encode), the goal is to create maximally different encodings (to be able to robustly distinguish them in the presence of noise or other ambiguities).
High-dimensional vector spaces offer plenty of space to push these vectors apart and moreover, they have the interesting property that \textit{random} vectors are already very far away \cite{Neubert2019}.
In particular for angular distance measures, this means that two random vectors are very likely almost orthogonal (this is called quasi-orthogonal): If we sample the direction of vectors independent and identically distributed (i.i.d.) from a uniform distribution, the more dimensions the vectors have, the higher is the probability that the angle between two such random vectors is close to 90 degrees; for 10,000 dimensional real vectors, the probability to be in \mbox{90$\pm$5} degrees is almost one. Please refer to \cite{Neubert2019} for a more in-depth presentation and evaluation.

The quasi-orthogonality property is heavily used in VSA operations. Since the different available VSAs use different vector spaces and metrics (cf. Sec.~\ref{subsec:sim_meas}), different approaches to create vectors are involved.
The most common approach is based on real numbers in the continuous range.
For instance, the Multiply-Add-Permute (\textbf{MAP-C} -- C stands for continuous) architecture uses the real range of $[-1,1]$. 
Other architectures such as \textbf{HRR}, \textbf{MBAT} as well as the \textbf{VTB} VSAs use a real range which is normally distributed with a mean of 0 and a variance of $1/D$ where $D$ defines the number of dimensions.
Another group uses binary vector spaces.
For example, the Binary Spatter Code (\textbf{BSC}) and the binary MAP (\textbf{MAP-B} as well as \textbf{MAP-I}) architecture generate the vectors in $\{0,1\}$ or $\{-1,1\}$. 
The creation of the binary values is based on a Bernoulli distribution with a probability of $p=0.5$. 
By reducing the probability $p$, sparse vectors can be created for the \textbf{BSDC-CDT}, \textbf{BSDC-S} as well as the \textbf{BSDC-SEG} VSAs (where the acronym CDT means Context Depend Thinning, S means shifting, and SEG means segmentally shifting, all three are binding operations and are explained in section \ref{subsec:binding}). 
To initialize the \textbf{BSDC-SEG} correctly, we use the density $p$ to calculate the number of segments $s=D \cdot p$ (this is needed for binding, as shown in Figure~\ref{img:segment_binding}) and randomly place a single 1 in each segment, all other entries are 0.
The authors of \cite{Rachkovskij2001} showed that a probability of $p=\frac{1}{\sqrt{D}}$ ($D$ is the number of dimensions) achieves the highest capacity\footnote{The term capacity refers to the number of stored items in the auto-associative memory in \cite{Rachkovskij2001}} in the vector and is therefore used in these architectures. 
Finally, a complex vector space can be used.
One example is the frequency Holographic Reduced Representations \textbf{FHRR} that uses complex numbers on the unit circle (the complex number in each vector dimension has length one) \cite{Plate94Phd}. 
It is therefore sufficient to use uniformly distributed values in the range of $(-\pi, \pi]$ to define the angles of the complex values -- thus, the complex vector can be stored using the real vector of angles $\theta$.
The complex numbers $c$ can be computed from the angles $\theta$ by $c=e^{i \cdot \theta}$.

\subsection{Similarity measurement}
\label{subsec:sim_meas}

VSAs use similarity metrics to evaluate vector representations, in particular, to find relations between two given vectors (figure out whether the represented symbols have a related meaning).
For example, given a noisy version of a hypervector as the output of a series of VSA operations, we might want to find the most similar elementary vector from a database of known symbols in order to decode this vector. 
A carefully chosen similarity metric is essential for finding the correct denoised vector from the database and to ensure a robust operation of VSAs.  
The term \textit{curse of dimensionality} \cite{Bellman61} describes the observation that algorithms that are designed for low dimensional spaces often fail in higher dimensional spaces -- this includes similarity measures based on Euclidean distance \cite{Beyer99}. 
Therefore, VSAs typically use other similarity metrics, usually based on angles between vectors or vector dimensions.

As shown in Table~\ref{tab:VSAs}, the architectures \textbf{MAP-C, MAP-B, MAP-I, HRR, MBAT} and \textbf{VTB} use the cosine similarity (cosine of the angle) between vectors $\mathbf{a}$ and $\mathbf{b} \in \mathbb{R}^D$: $s = sim(\mathbf{a},  \mathbf{b})=cos(\mathbf{a}, \mathbf{b})$. 
The output is a scalar value ($\mathbb{R}^{D} \times \mathbb{R}^{D} \longrightarrow \mathbb{R}$) within the range $[-1,1]$.
Note that -1 means collinear vectors in opposite directions and 1 means identical directions.
A value of 0 indicates orthogonal vectors.

The binary vector space can be combined with different similarity metrics depending on the sparsity:
Either the complementary Hamming Distance for binary dense vectors, like \textbf{BSC} or the overlap for binary sparse vectors as \textbf{BSDC-CDT, BSDC-S, BSDC-SEG} (the overlap can be normalized to the range $[0,1]$ (0 means non-similar and 1 means similar)). 
Eq.~\ref{eq:bin_sim} shows the equation to compute the similarity (complementary and normalized Hamming Distance) between dense ($p=0.5$) binary vectors (\textbf{BSC}) $\mathbf{a}$ and $\mathbf{b} \in \{0,1\}^D$, given the number of dimensions $D$. 

\begin{equation}
s = sim(\mathbf{a},  \mathbf{b})= 1- {\frac{HammingDist(\mathbf{a},   \mathbf{b})}{D}}
\label{eq:bin_sim}
\end{equation}

The complex space needs yet another similarity measurement.
As introduced in section \ref{subsec:hv_elemts}, the complex architecture of \cite{Plate94Phd} (\textbf{FHRR}) uses angles $\theta$ of complex numbers. 
To measure how similar two vectors are, the average angular distance is calculated (keep in mind, since the complex vectors have unit length, vectors $\mathbf{a}$ and  $\mathbf{b}$ are from $\mathbb{R}^D$ and only contain the angles $\theta$): 
\begin{equation}
s = sim(\mathbf{a},  \mathbf{b})= \frac{1}{D} \cdot \sum _{i=1}^D cos(a_i-b_i)
\end{equation}

\subsection{Bundling}
\label{subsec:bundling}

VSAs use the bundling operator to superimpose (or overlay) given hypervectors (similar to what was done in the introductory example).
Bundling aggregates a set of input vectors of space $\mathbb{V}$ and creates an output vector of the same space $\mathbb{V}$ that is similar to its inputs.
Plate \cite{Plate1997} declared that the essential property of the bundling operator is the unstructured similarity preservation. 
It means: a bundle of vectors \textbf{A} + \textbf{B} is still similar to vector \textbf{A}, \textbf{B} and also to another bundle \textbf{A} + \textbf{C} that contains one of the input vectors. Since all compared VSAs implement bundling as an addition-like operator, the most commonly used symbol for the bundling operation is $+$. 

The implementation is typically a simple element-wise addition.
Depending on the vector space it is followed by a normalization step to the specific numerical range.
For instance vectors of the \textbf{HRR}, \textbf{VTB} and  \textbf{MBAT} have to be scaled to a vector length of one. 
Bundled vectors from the \textbf{MAP-C} are cut at -1 and 1.
The binary VSAs \textbf{BSC} and \textbf{MAP-B} use a threshold to convert the sums into the binary range of values. 
The threshold depends on the number of bundled vectors and is exactly half this number. 
Potential ties in case of an even number of bundled vectors are decided randomly. 
In the sparse distributed architectures, the logical OR function is used to implement the bundling operation. Since only a few values are non-zero, they carry most information and shall be preserved. 
For example, \cite{Rachkovskij2001} do not apply thinning after bundling, however, in some application it is necessary to decrease the density of the bundled vector. 
For instance, the language recognition example in section \ref{subsec:lang_rec} requires a density constraint -- we used a (empirically determined) maximum density of 50\%. 
Besides the \textbf{BSDC} without thinning, the \textbf{MAP-I} does not need normalization as well -- it accumulates the vectors withing the integer range.
The bundling operator in \textbf{FHRR} first converts the angle vectors to the form $e^{i \cdot \theta}$ before using element-wise addition. 
Afterward, the complex-valued vectors will be added. 
Then, only the angles of the resulting complex numbers are used and the magnitudes are discarded -- the output are the new angles $\theta$. 
The complete bundling step is shown in equation \ref{eq:bundling_FHRR}:

\begin{equation}
\mathbf{a} + \mathbf{b} = angle(e^{i \cdot a} + e^{i \cdot b})
\label{eq:bundling_FHRR}
\end{equation}

Due to its implementation in form of addition, bundling is commutative and associative in all compared VSA implementations except for the normalized bundling operations which are only approximately associative: $(\mathbf{A}+\mathbf{B}) + \mathbf{C} \approx \mathbf{A} +(\mathbf{B} + \mathbf{C})$.

\subsection{Binding}
\label{subsec:binding}

The binding operator is used to connect two vectors, e.g., the role-filler pairs in the introduction. The output is again a vector from the same vector space. Typically, it is the most complex and most diverse operator of VSAs. 
Plate \cite{Plate1997} defines the properties of the binding as follows: 
\begin{itemize}
\item the output is non-similar to the inputs: binding of \textbf{A} and \textbf{B} is non similar to \textbf{A} and \textbf{B}
\item it preserves structured similarity: binding of \textbf{A} and \textbf{B} is similar to binding of \textbf{A'} and \textbf{B'}, if  \textbf{A'} is similar to \textbf{A} and \textbf{B'} is similar to \textbf{B}
\item an inverse of the operation exists (defined as unbinding with symbol $\oslash$)
\end{itemize}
The binding  is typically indicated by the mathematical symbol $\otimes$.

Unbinding $\oslash$ is required to recover the elemental vectors from the result of a binding \cite{Plate1997}.
Given a binding \textbf{C}=\textbf{A}$\otimes$\textbf{B}, we can retrieve the elemental vectors \textbf{A} or \textbf{B} from \textbf{C} with the unbinding operator: \textbf{R}=\textbf{A}$\oslash$\textbf{C}  (or \textbf{B}$\oslash$\textbf{C}).
\textbf{R} is now similar to the vector \textbf{B} or \textbf{A} respectively.

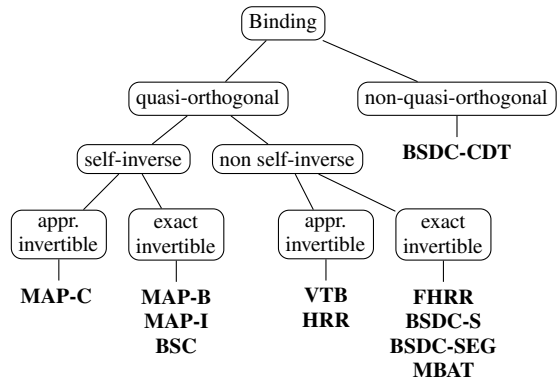
\begin{figure}[t]
\begin{center}
\begin{tikzpicture}[sibling distance=7em,
  every node/.style = {shape=rectangle, rounded corners,
    draw, align=center, 
    top color=white}]]
  \node (bind)[align=center]{Binding}
    child { node(mul)[yshift=0.5cm]{quasi-orthogonal}
      child { node(se)[yshift=0.7cm]{self-inverse}
        child { node(app1)[, yshift=0.5cm]{appr. \\ invertible} 
                child { node[draw=none, below=1em of app1]{\textbf{MAP-C} } }
                }
        child { node(ex1)[right=1em of app1] {exact \\ invertible} 
                child { node(mapd)[draw=none, below=1em of ex1]{\textbf{MAP-B} \\ \textbf{MAP-I} \\ \textbf{BSC}}}}
        }        
      child { node(nse)[right=1em of se]{non self-inverse}  
          child { node(app2)[right=8em of app1]{appr. \\ invertible} 
              child { node(mapd)[draw=none, below=1em of app2]{\textbf{VTB} \\ \textbf{HRR} }}        
              }
          child { node(ex2)[right=1em of app2]{exact \\ invertible}
              child { node(mapd)[draw=none, below=1em of ex2]{\textbf{FHRR} \\ \textbf{BSDC-S} \\ \textbf{BSDC-SEG} \\ \textbf{MBAT}}}
              }
              }    
          }
      child { node(add)[xshift=1.3cm,yshift=0.5cm]{non-quasi-orthogonal}
            child { node[draw=none, below=1em of add]{\textbf{BSDC-CDT}} }
            }
          ;
\end{tikzpicture}
\end{center}
\caption{Taxonomy of different binding operations. The VSAs that use each binding are printed in bold (see the Table~\ref{tab:VSAs} for more details).}
\label{fig: taxonomy}
\end{figure}

From a historical perspective, one of the first ideas to associate connectionist representations goes back to Smolensky \cite{smolensky90}. 
He uses the tensor product (the outer product of given vectors) to compute a representation that combines all information of the inputs. 
To recover (unbind) the input information from the created matrix, it requires only the normalized inner product of the vector with the matrix (the tensor product). 
Based on this procedure, it is possible to perform exact binding and unbinding (recovering). 
However, using the tensor product creates a problem: the output of the tensor product of two vectors is a matrix and the size of the representation grows with each level of computation. 
Therefore, it is preferable to have binding operations (and corresponding unbinding operations) that \textit{approximate} the result of the outer product in a vector ($\mathbb{V} \times \mathbb{V} \rightarrow \mathbb{V}$). 
Thus, according to Gayler \cite{gayler03} a VSA's binding operation is basically a tensor product representation followed by a function to preserve the dimensionality of the input vectors. 
For instance, \cite{Frady2021} shows that the Hadamard product in the MAP VSA is a function of the outer product.
Based on this dimensionality preserving definition, several binding and unbinding operations have been developed specifically for each vector domain.
These different binding operations can be arranged in the taxonomy shown in Fig.~\ref{fig: taxonomy}.

The existing binding implementations can be basically divided into two types: quasi-orthogonal and non-quasi-orthogonal (see Fig.~\ref{fig: taxonomy}). 
Quasi-orthogonal bindings explicitly follow the properties of Plate \cite{Plate1997} and generate an output that is dissimilar to their inputs. 
In contrast, the output of a non-quasi-orthogonal binding will be similar to the input.
Such a binding operation requires additional computational steps to achieve the properties specified by Plate (for example a nearest-neighbor search in an item memory \cite{Rachkovskij2001}).

On the next level of the taxonomy, quasi-orthogonal bindings can be further distinguished into self-inverse and non self-inverse binding operations. 
Self-inverse refers to the property that the inverse of the binding is the binding operation itself (unbinding=binding)\footnote{It should be noted that the operator is commonly referred to as self-inverse, but it is rather the vector that has this property and not the operator.}.
The opposite is the non self-inverse binding: it requires an additional unbinding operator (inverse of the binding). 
Finally, each of these nodes can be separated into approximate and exact invertible binding (unbinding). 
For instance, the Smolensky tensor product is an exact invertible binding, because the unbinding produces exactly the same vector as in the input of the binding: $\mathbf{a} \oslash (\mathbf{a} \otimes \mathbf{b}) = \mathbf{b} $.
The approximate inverse produces an unbinding output which is similar to the input of the binding, but not the same: $\mathbf{a} \oslash (\mathbf{a} \otimes \mathbf{b}) \approx \mathbf{b}$.

An quasi-orthogonal binding can be, for example, implemented by \textit{element-wise multiplication} (as in \cite{Gayler98}).
In case of bipolar values ($\pm 1$), element-wise multiplication is self-inverse, since $1^2=-1^2=1$.
The self-inverse property is essential for some VSA algorithms in the field of analogical reasoning (this will be the topic of Sec.~\ref{subsec:analog}).
Element-wise multiplication is, for example, used in the \textbf{MAP-C}, \textbf{MAP-B} and \textbf{MAP-I} architectures.
An important difference is that for the continuous space of \textbf{MAP-C} the unbinding is only approximate while it is exact for the binary space in \textbf{MAP-B}. 
For \textbf{MAP-I} it is exact for elementary vectors (from $\{-1,1\}$) and approximate for processed vectors. 
Compared to the Smolensky tensor product, element-wise multiplication approximates the outer product matrix by its diagonal. 
Further, the element-wise multiplication is both commutative and associative (cf. Table~\ref{tab:VSAs}). 

Another self-inverse binding with an exact inverse is defined in the \textbf{BSC} architecture. 
It uses the \textit{exclusive or (XOR)} and is equivalent to the element-wise multiplication in the bipolar space.
As expected, the XOR is used for both binding and unbinding -- it provides an exact inverse.  
Additionally, it is commutative and associative like element-wise multiplication.

The second category within the quasi-orthogonal bindings in our taxonomy in Fig.~\ref{fig: taxonomy} are non self-inverse bindings.  
Two VSAs have an approximate unbinding operator. 
Binding of the real-valued vectors of the \textbf{VTB} architecture are computed using \textit{Vector Derived Transformation (VTB)} as described in \cite{Gosmann2019}. 
They use a matrix multiplication for binding and unbinding. 
The matrix is constructed from the second input vector $\mathbf{b}$, and multiplied with the first vector $\mathbf{a}$ afterward. 
Eq.~\ref{eq:vtb} formulates the VTB as binding where $V_b^\prime$ represents a square matrix (equation \ref{eq:matrix_v}) which is the reshaped vector $b$.

\begin{equation}
\mathbf{c}=\mathbf{a} \otimes \mathbf{b} = V_{b} \cdot \mathbf{a}=\left[\begin{array}{ccc}{V_{b}^{\prime}} & {0} & {0} \\ {0} & {V_{b}^{\prime}} & {0} \\ {0} & {0} & {\ddots}\end{array}\right] \mathbf{a}
\label{eq:vtb}
\end{equation}

\begin{equation}
 V_{b}^{\prime}=d^{\frac{1}{4}}\left[\begin{array}{cccc}{b_{1}} & {b_{2}} & {\cdots} & {b_{d^{\prime}}} \\ {b_{d^{\prime}+1}} & {b_{d^{\prime}+2}} & {\cdots} & {b_{2 d^{\prime}}} \\ {\vdots} & {\vdots} & {\ddots} & {\vdots} \\ {b_{d-d^{\prime}+1}} & {b_{d-d^{\prime}+2}} & {\cdots} & {b_{d}}\end{array}\right], d^{\prime}=\sqrt{D}
\label{eq:matrix_v}
\end{equation}

This specifically designed transformation matrix (based on the second vector) provides a stringent transformation of the first vector which is invertible (i.e. it allows unbinding). 
This unbinding operator is identical to binding in terms of matrix multiplication, but the transposed matrix $V_b$ is used for calculation, as shown in the eq.~\ref{eq:vtb_unbind}. 
These binding and bundling operations are neither commutative nor associative. 

\begin{equation}
\mathbf{a} \approx \mathbf{b} \oslash \mathbf{c} = V_{b}^{\top} \mathbf{c}
\label{eq:vtb_unbind}
\end{equation}

Another approximated non self-invertible binding is part of the \textbf{HRR} architecture: the \textit{circular convolution}.
Binding of two vectors $\mathbf{a}$ and $\mathbf{b} \in \mathbb{R}^D$ with circular convolution is calculated by:

\begin{equation}
\mathbf{c}=\mathbf{a} \otimes \mathbf{b}\hspace{-0.1cm}:  c_{j}=\sum_{k=0}^{D-1} b_{k} a_{mod(j-k,D)}
\hspace{0.1cm} \textrm{with} \hspace{0.2cm} j\in \{0,...,D-1\}
\label{eq:circ_conv}
\end{equation}

Circular convolution approximates Smolensky's outer product matrix by sums over all of its (wrap-around) diagonals. For more details pleaser refer to \cite{Plate1995}. 
Based on the algebraic properties of convolution, this operator is commutative as well as associative. 
However, convolution is not self-inverse and requires a specific unbinding operator.
The circular correlation (eq.~\ref{eq:circ_corr}) provides an approximated inverse of the circular convolution and is used for unbinding.
It is neither commutative nor associative.

\begin{equation}
\mathbf{a}\approx \mathbf{b} \oslash \mathbf{c}\hspace{-0.1cm}:   a_{j}=\sum_{k=0}^{D-1} b_{k} c_{mod(k+j,D)}
\hspace{0.1cm} \textrm{with} \hspace{0.2cm} j\in \{0,...,D-1\}
\label{eq:circ_corr}
\end{equation}

A useful property of the convolution is that it becomes an element-wise multiplication in the frequency domain (complex space). 
Thus, it is possible to operate entirely in the complex vector space and use the element-wise multiplication as the binding operator \cite{Plate94Phd}. 
This leads to the \textbf{FHRR} VSA with an exact invertible and non self-inverse binding\footnote{It should be noted that there are relations between operations of different VSAs and between self-inverse and non self-inverse bindings: If the angles of an FHRR are quantized to two levels (e.g., $\{0, \pi \}$), the binding becomes self-inverse and equivalent to binary VSAs like BSC or MAP-B.} as shown in the taxonomy in Fig.~\ref{fig: taxonomy}. 
With the constraints described in Sec.~\ref{subsec:hv_elemts} (using complex values with a length of one), the computation of binding and unbinding becomes more efficient.
Given two complex numbers \textit{$c_1$} and \textit{$c_2$} with angles $\theta _1$ and $\theta _2$ and length 1, multiplication of the complex numbers becomes an \textit{addition} of the angles:

\begin{equation}
c_1 \cdot c_2 = e^{i\cdot \theta_1} \cdot e^{i\cdot \theta_2} = e^{i \cdot (\theta_1 + \theta_2)}
\label{eq:complex_mul}
\end{equation}

The same procedure applies to unbinding but with the angles of the conjugates of one of the given vectors -- hence, it is just a \textit{subtraction} of the angles $\theta_1$ and $\theta_2$. 
Note that a modulo operation with $2 \pi$ (angles on the complex plane are in the range of $(-\pi,\pi]$) must follow the addition or subtraction.  
Based on this assumption, it is possible to operate only with the angles rather than the whole complex numbers. 
Since the addition is associative and commutative, the binding is as well. 
But analog to the unbinding operation, subtraction is non-commutative and non-associative -- therefore is also the unbinding. 
At this point we would like to emphasize that \textbf{HRR} and \textbf{FHRR} are basically functionally equivalent -- the operations are performed either in spatial or frequency domain. 
However, the assumption of unit magnitudes in \textbf{FHRR} distinguishes both and simplifies the implementation of the binding.
Moreover, in contrast to \textbf{FHRR}, \textbf{HRR} uses an approximate unbinding because it is more stable and robust against noise compared to an exact inverse \cite[p.102]{Plate94Phd}.

In the following, we describe the two sparse VSAs with an quasi-orthogonal, exact invertible and non self-inverse binding: the \textbf{BSDC-S} (binary sparse distributed representations with shifting) and the \textbf{BSDC-SEG} (sparse vectors with segmental shifting as in \cite{Laiho2015}).
The shifting operation allows to encode hypervectors into a new representation which is dissimilar to the input. 
Either the entire vector is shifted by a certain number or divided into segments and each segment is shifted individually by different values. 
The former goes as follows: Given two vectors, the first will be converted to a single hash-value (e.g. use the on-bits' position indices). 
Afterwards, the second vector is shifted by this hash-value (circular shifting).
This operation has an exact inverse (shifting in the opposite direction), but it is neither commutative nor associative.

\begin{figure}[thb]
\centering
\includegraphics[width=\textwidth]{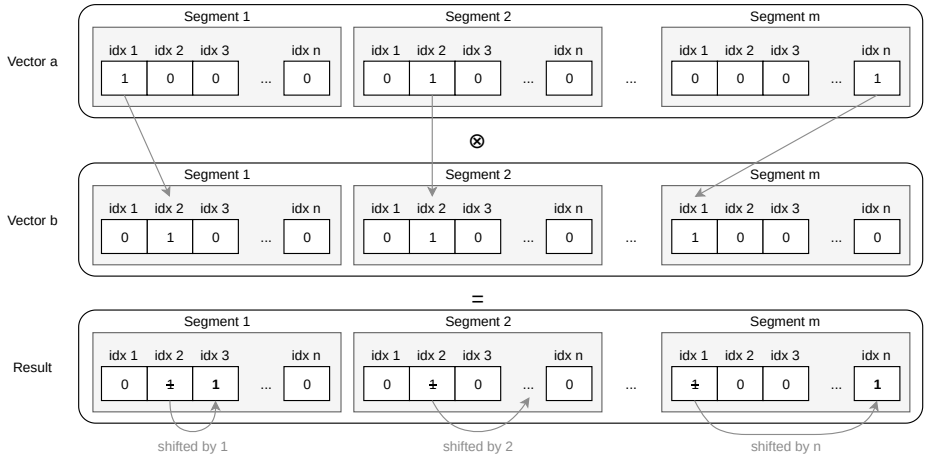}
\caption{Segment-wise shifting for binding sparse binary vectors $a$ and $b$.}
\label{img:segment_binding}
\end{figure}

The latter (segment-wise shifting -- \textbf{BSDC-SEG}) includes additional computing steps: 
As described in \cite{Laiho2015}, the vectors are split into segments of the same length. 
Preferably, the number of segments depends on the density and is equal to the number of on-bits in the vector -- thus, we have one on-bit per segment in average. 
For better understanding, see Fig.~\ref{img:segment_binding} for binding vector $a$ with vector $b$. 
Each of those vectors has $m$ segments (gray shaded boxes) with $n$ values (bits). 
The position of the first on-bit in each segment of the vector gives one index per segment. 
Next, the segments of the second vector $b$ will be circularly shifted by these indices (see the resulting vector in the figure). 
Like the BSDC-S, the unbinding is just a simple shifting by the negated indices of the vector $ a $.
Since the binding of this VSA resembles an addition of the segment indices, it is both commutative and associative.
In contrast, the unbinding operation is a subtraction of the indices of vector a and b and is neither commutative nor associative. 
As mentioned earlier, different binding operations can be related. As another example, the binding operation of \textbf{BSDC-SEG} corresponds to an angular representation as in \textbf{FHRR} with $m$ elements quantized to $n$ levels.

The last VSA with an exact invertible binding mechanism is \textbf{MBAT}. 
It is similar to the earlier mentioned VTB binding that constructs a matrix to bind two vectors. 
MBAT \cite{Gallant2013} uses matrices with a size of $D \times D $ to bind vectors of length $D$ -- this procedure is similar to the Smolenskys tensor product. 
The binding matrix must be orthonormal and can be transposed to unbind a vector. 
To avoid creating a completely new matrix for each binding, \cite{Tissera2014} uses an initial orthonormal matrix $ M $ and manipulates it for each binding. 
It uses the exponentiation of the initial matrix $ M $ by an arbitrary index $i$, resulting in a matrix $M_i$ that is still orthonormal but after binding gives a different result than the initial matrix $M$. 
For our experimental comparison, we randomly sampled the initial matrix from an uniform distribution and convert it to an orthonormal matrix with the singular value decomposition.
Since exponentiation of the initial matrix $M$ leads to a high computational effort, we approximate the matrix manipulation by shifting the rows and the columns by the appropriate index of the role vector.  
This index is calculated with a hash-value of the role vector (simple summation over all indices of elements greater than zero).
However, like the \textbf{VTB} VSA, the \textbf{MBAT} binding and unbinding are neither commutative nor associative.

According to Fig.~\ref{fig: taxonomy}, there is one VSA that uses a non-orthogonal binding. 
The \textbf{BSDC-CDT} from Rachkovskij et al. \cite{Rachkovskij2001} introduces a binding operator for sparse binary vectors with an additive operator: the disjunction (logical OR). 
Since disjunction of sparse vectors can produce up to twice the number of on bits, they propose a \textit{Context Depend Thinning (CDT)} procedure to thin vectors after the disjunction. 
The complete CDT procedure is described in \cite{Rachkovskij2001a}. 
Since this binding operation creates an output that is similar to the inputs, it is in contrast to Plate's \cite{Plate1997} properties of binding operators (from the beginning of this section). 
As a consequence, instead of using unbinding to retrieve elemental vectors, the similarity to all elemental vectors has to be used to find the most similar ones.
In contrast to the previously discussed quasi-orthogonal binding operations, here, additional computational steps are required to achieve the properties of the binding procedure defined by Plate \cite{Plate1997}.
Particularly, if the CDT is used for consecutive binding and bundling (e.g., bundling role-filler pairs can be seen as two levels -- first is binding and second is bundling), this requires to store the specific level (binding at first level and bundling at the second level). 
During retrieval, the similarity search (unbinding) must be done in the corresponding level of binding, because this binding operator preserves the similarity of all bound vectors (in this example, every elemental vector is similar to the final representation after binding and bundling). 
Based on such iterative search (from level to level), the CDT binding needs more computational steps and is not directly comparable with the other binding operations. 
Therefore, the later experimental evaluations will use the segment-wise shifting as binding and unbinding for both the BSDC-S and BSDC-SEG VSAs instead of the CDT.

Finally, we want to emphasize the different complexities of the binding operations. 
Based on a comparison in \cite{Kelly2013}, for $D$ dimensional vectors, the complexities (number of computing steps) of binding two vectors are as follows: \\

\begin{tabular}{lll}
- & element-wise multiplication & \\
  & (MAP-C, MAP-B, BSC, FHRR): &  $O(D)$  \\
- & circular conv. (HRR): & $O(D \ log \ D)$ \\
- & matrix binding (MBAT, VTB): & $O(D^2)$ \\
- & sparse shifting (BSDC-S, BSDC-SEG)\footnotemark: & $O(D)$ \\
\end{tabular} \\

\footnotetext{Number of computational steps also depends on the density $p$.}

\subsection{Ramifications of non self-inverse binding}
\label{subsec:analog}

Sec.~\ref{subsec:binding} distinguished two different types of binding operations: self-inverse and non self-inverse. 
We want to demonstrate possible ramifications of this property using the classical example from Pentti Kanerva on analogical reasoning \cite{Kanerva2010}: "What is the Dollar of Mexico?"
The task is as follows:
Similar to the representation of the country $USA$ ($R_{USA} = Name \otimes USA + Curr \otimes Dollar + Cap \otimes WDC$) from the example in the introduction, we can define a second representation of the country $Mexico$: 
\begin{equation}
R_{Mex}=Name \otimes Mex + Curr \otimes Peso + Cap \otimes MXC
\label{eq:rep_mexico}
\end{equation}
Given these two representations, we, as humans, can answer Kanerva's question by analogical reasoning: Dollar is the currency of the USA, the currency of Mexico is Peso, thus the answer to the above question is ``Peso''.
This procedure can be elegantly implemented using a VSA. However, the method described in \cite{Kanerva2010} only works with self-inverse bindings, such as \textbf{BSC} and \textbf{MAP}.
To understand why, we will explain the VSA approach more in detail:
Given are the records of both countries $R_{Mex}$ and $R_{USA}$ (the latter is written out in the introduction).
In order to evaluate analogies between these two countries, we can combine all the information from these two representations into a single vector using binding.
This creates a mapping $F$: 
\begin{equation}
F=R_{USA} \otimes R_{Mex}
\label{eq:F}
\end{equation}
With the resulting vector representation we can answer the initial question ("What is the Dollar of Mexico?") by binding the query vector (Dollar) to the mapping: 
\begin{equation}
A=Dol \otimes F \approx Peso
\label{eq:F_Dol}
\end{equation}
The following explains why this actually works.
Eq.~\ref{eq:F} can be examined based on the algebraic properties of the binding and bundling operations (e.g. binding distributes over bundling). 
In case of a self-inverse binding (cf. taxonomy in Fig.~\ref{fig: taxonomy}), the following terms result from eq. \ref{eq:F} (we refer to Kanerva \cite{Kanerva2010} for a more detailed explanation): 
\begin{equation}
F=(USA \otimes Mex) + (Dol \otimes Peso) + (WDC  \otimes MXC) + N
\label{eq:F_important}
\end{equation}
Based on the self-inverse property, terms like $Curr \otimes Curr$ cancel out (i.e. they create a ones-vector). 
Since binding creates an output that is not similar to the inputs, other terms, like $Name \otimes Curr$, can be treated as noise and they are summarized in the term $N$.
The noise terms are dissimilar to all known vectors and basically behave like random vectors (which are quasi-orthogonal in high-dimensional spaces). 
Binding the vector $Dol$ to the mapping $F$ of USA and Mexico (eq.~\ref{eq:F_Dol}) creates vector $A$ in eq.~\ref{eq:F_Dol_det} (only the most important terms are shown).
The part $Dol \otimes (Dol \otimes Peso)$ is important because it reduces to $Peso$, again, based on the self-inverse property. 
As before, the remaining terms behave like noise that is bundled with the representation of $Peso$. 
Since the elemental vectors (representations for, e.g., $Dollar$ or $Peso$) are randomly generated, they are highly robust against noise. 
That is why the resulting vector $A$ is still very similar to the elemental vector for $Peso$. 
\begin{equation}
A=Dol \otimes ((USA \otimes Mex) + (Dol \otimes Peso) + ...+ N)
\label{eq:F_Dol_det}
\end{equation}
Notice, the previous description is only a brief summary to the ``Dollar of Mexico'' example. 
We refer to \cite{Kanerva2010} for more details. 

However, we can see that the computation is based on a self-inverse binding operation. 
As described in Sec.~\ref{sec:vsa_props} and the taxonomy in Fig.~\ref{fig: taxonomy}, some VSAs have no self-inverse binding and need an unbind operator to retrieve elemental vectors.

The above described approach \cite{Kanerva2010} has the particularly elegant property that all information about the two records is stored in the single vector $F$ and once this vector is computed, any number of queries can be done, each with a single operation (eq.~\ref{eq:F_Dol}).
However, if we relax this requirement, we can address the same task with the two-step approach described in \cite[p.~265]{Kanerva2001}. 
This also relaxes the requirement of a self-inverse binding and uses unbinding instead:
\begin{equation}
A = R_{Mex} \oslash (R_{USA} \oslash Dol) \\
\label{eq:A_unbinding}
\end{equation}
After simplification to the necessary terms (all other terms are represented as noise $N$), we get equation~\ref{eq:A_unbinding_simpl}.
\begin{align}
A &= (\underbrace{Curr}_{Role} \otimes \underbrace{Peso}_{Filler}) \oslash ((\underbrace{Curr}_{Role} \otimes \underbrace{Dol}_{Filler}) \oslash \underbrace{Dol}_{Filler}) + N \label{eq:A_unbinding_simpl}  \\
A &= (\underbrace{Curr}_{Role} \otimes \underbrace{Peso}_{Filler}) \oslash \underbrace{Curr}_{Role} +N \nonumber \\
A &= Peso + N \nonumber
\end{align}
It can be seen that it is in principle possible to solve the task 'What is the dollar of Mexico?' with non-self-inverse binding operators. 
However, this requires storing more vectors (both $R_{Mex}$ and $R_{USA}$ are stored) and additional computational effort.

In the same direction, Plate \cite{Plate1995} emphasized the need for a 'readout' machine for the HRR VSA to decode chunked sequences (hierarchical binding). 
It retrieves the trace iteratively and finally generates the result. 
Transferred to the given example: first, we have to figure out the meaning of \textit{Dollar} (it is the currency of the USA) and query the result (\textit{Currency}) on the representation of Mexico afterward (resulting in \textit{Peso}). 
Such a readout requires more computation steps caused by iteratively traversing of the hierarchy tree (please see \cite{Plate1995} for more details).
Presumably, this is a general problem of all non self-inverse binding operations.

\section{Experimental Comparison}
\label{sec:exp}

After the discussion of theoretical aspects in the previous section, this section provides an experimental comparison of the different VSA implementations using three experiments. The first evaluates the bundling operations to answer the question
\textit{How efficiently can the different VSAs store (bundle) information into one representation?}
The topic of the second experiment are the binding and unbinding operations.
As described in Sec.~\ref{subsec:binding} and the taxonomy in Fig.~\ref{fig: taxonomy}, some binding operations have an approximate inverse. 
Hence, the second experiment evaluates the question
\textit{How good is the approximation of the binding inverse?}
Finally, the third experiment focuses on the combination of bundling and binding and the ability to recover noisy representations. 
There, the leading question is: 
\textit{To what extent are binding and unbinding disturbed by bundled representations?}  

\textbf{A note on the evaluation setup}
We will base our evaluation on the required \textit{number of dimensions} of a VSA to achieve a certain performance instead of the physical \textit{memory consumption} or \textit{computational effort} - although the storage size and the computational effort per dimension can vary significantly (e.g. between a binary vector and a float vector). 
The main reason is that the actual resource demands of a single VSA might vary significantly dependent on the capabilities and limitations of the underlying hard- and software, as well as the current task.
For example, it is well-known that HRR representations do not require a high precision for many tasks \cite[p. 67]{Plate94Phd}. However, low resolution data types (e.g. half-precision floats or less) might not be available in the used programming language. Instead, using the number of dimensions introduces a bias towards VSAs with high memory requirements per dimension, however, the values are supposed to be simple to convert to actual demands given a particular application setup.

\subsection{Bundling Capacity}
\label{subsec:bundle_cap}

We evaluate the question \textit{How efficiently can the different VSAs store (bundle) information into one representation?}
We use an experimental setup similar to \cite{Neubert2019}, extend it with varying dataset sizes and varying numbers of dimensions, and use it to experimentally compare the eleven VSAs. 
For each VSA, we create a database of $N=1,000$ random elementary vectors from the underlying vector space $\mathbb{V}$. 
It represents basic entities stored in a so-called item memory.
To evaluate the bundle capacity of this VSA, we randomly chose $k$ elementary vectors (without replacement) from this database and create their superposition $B \in \mathbb{V}$ using the VSA's bundle operator.
Now the question is whether this combined vector $B$ is still similar to the bundled elementary vectors.
To answer this question, we query the database with the vector $B$ to obtain the k elementary vectors, which are the most similar to the bundle $B$ (using the VSA's similarity metric). 
The evaluation criterion is the accuracy of the query result: the ratio of correctly retrieved elementary vectors on the k returned vectors from the database.\footnote{This experimental setup is closely related to Bloom filters that can efficiently evaluate whether an element is part of a set. Their relation to VSAs is discussed in \cite{Kleyko2020}.}


The capacity depends on the dimensionality of $\mathbb{V}$. Therefore we range the number of dimensions $D$ in 4...1156 (since \textbf{VTB} needs even roots the number of dimensions is computed by $i^2$ with $i=2 ... 34$) and evaluate for $k$ in 2...50. We use $N=1,000$ elementary vectors. 
To account for randomness, we repeat each experiment 10 times and report means.

\begin{figure}[!t]
\centering
\includegraphics[width=\textwidth]{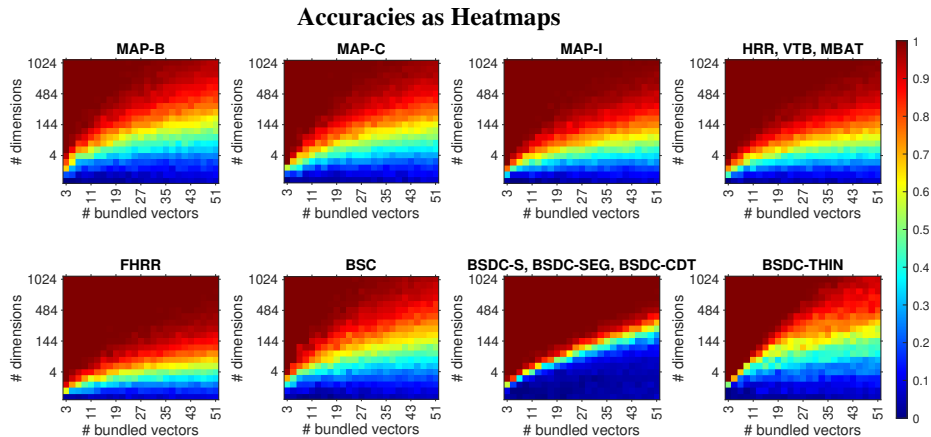}
\caption{Heat-maps showing the accuracies of different number of bundled vectors and numbers of dimensions.}
\label{img:bundle_results_2d}
\end{figure}

Fig.~\ref{img:bundle_results_2d} shows the results of the experiment in form of a heat-map for each VSA, which encodes the accuracies of all combinations of number of bundled vectors and number of dimensions in colors.
The warmer the color, the higher the achieved accuracy with a particular number of dimensions to store and retrieve a certain number of bundled vectors. 
One important observation is the large dark red areas (close to perfect accuracies) achieved by the \textbf{FHRR} and \textbf{BSDC} architectures. 
Also remarkable is the fast transition from very low accuracy (blue) to perfect accuracy (dark red) for the \textbf{BSDC} architectures; dependent on the number of dimensions, bundling will either fail or work almost perfectly.
Presumably, this is the result of the increased density after bundling without thinning.
The last plot in Fig.~\ref{img:bundle_results_2d} shows how the transition range between low and high accuracies increases when using an additional thinning (with maximum density 0.5)\footnote{Since the BSDC architecture performance also depends on the given sparsity, we want to refer to \cite{Kleyko2018} for a more exhaustive sensitivity analysis of sparse vectors on a classification task.}.

\begin{figure}[!t]
\centering
\includegraphics[width=\textwidth]{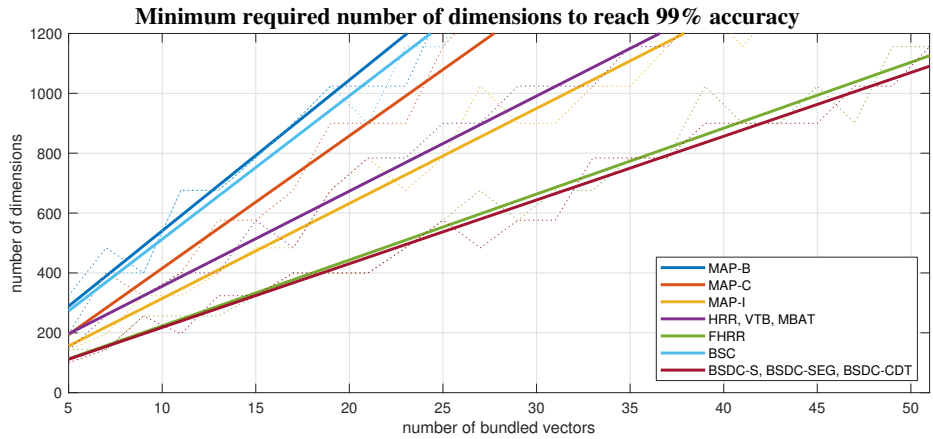}
\caption{Minimum required number of dimensions to reach 99\% accuracy. The solid lines represent linear fitted curves. The flatter the curves/lines, the more efficient is the bundling. Keep in mind, different VSAs might have very different memory consumption per dimension.}
\label{img:bundle_results}
\end{figure}

For an easier access to the different VSAs performances in the capacity experiment, Fig.~\ref{img:bundle_results} summarizes the results of the heatmaps in 1-D curves. 
It provides an evaluation of the required number of dimensions to achieve almost perfect retrieval for different values of k.
We selected a threshold of 99\% accuracy, that means 99 of 100 query results are correct. 
A threshold of 100\% would have been particularly sensitive to outliers, since a single wrong retrieval would prevent achieving the 100\%, independent of the number of perfect retrieval cases.
To make the comparison more accessible, we fit a straight line to the data points and plot the result as a dotted line. %

Dense binary spaces need the highest number of dimensions, real-valued vectors a little less and the complex values require the smallest number of dimensions. 
As expected from the previous plots in Fig.~ \ref{img:bundle_results_2d}, the binary sparse (\textbf{BSDC, BSDC-S, BSDC-SEG}) and the complex domain (\textbf{FHRR}) reach the most efficient results. 
They need fewer dimensions to bundle all vectors correctly. 
The sparse binary representations perform better than the dense binary vectors in this experiment. A more in-depth analysis of the general benefits of sparse distributed representations can be found in \cite{Ahmad2019}.
Particularly interesting is also the comparison between the \textbf{HRR} VSA from \cite{Plate1995} and the complex-valued \textbf{FHRR} VSA from \cite{Plate94Phd}. 
Both the \textbf{FHRR} with the complex domain as well as the \textbf{HRR} architecture operate in a continuous space (where values in \textbf{FHRR} represent angles of unit-length complex numbers). 
However, operating with real values in a complex perspective increases the efficiency noticeably. 
Even if the \textbf{HRR} architecture is adapted to a range of $[-\pi, \pi]$ like the complex domain, the performance of the real VSA does not change remarkably. 
This is an interesting insight: If real numbers are treated as if they were angles of a complex number, then this increases the efficiency of bundling. 

We want to emphasize again that different VSAs potentially require very different amounts of memory per dimension.
Very interestingly, in these experiments, the sparse vectors require a low number of dimensions and are additionally expected to have particularly low memory consumption.
A more in-depth evaluation of memory and computational demands is an important point for future work.

Besides the experimental evaluation of the bundle capacity, the literature provides analytical methods to predict the accuracy for a given number of bundled vectors and number of dimensions. 
Since it this is not yet available for all of our evaluated VSAs, we have not used it in our comparison. 
However, we found a high accordance of our experimental results with the available analytical results. 
Further information about analytical capacity calculation can be found in \cite{Gallant2013}, \cite{Frady2018} and \cite{Kleyko2018a}. 

\textbf{Influence of the item memory size}
In the above experiments, we used a fixed number of vectors in the item memory ($N=1,000$).
Plate \cite[p. 160 ff]{Plate94Phd} describes a dependency between the size of the item memory and the accuracy of the superposition memory (bundled vectors) for Holographic Reduced Representations.
The conclusion was that the number of vectors in the item memory ($N$) can be increased exponentially in the number of dimensions $D$ while maintaining the retrieval accuracy. 
To evaluate the influence of the item memory size for all VSAs, we slightly modify our previous experimental setup. This time, we fix the number of bundled vectors to $k=10$ and report the minimum number of dimensions that is required to achieve an accuracy of at least 99\% for a varying number $N$ of elements in the item memory.

\begin{figure}[!t]
\centering
\includegraphics[width=\textwidth]{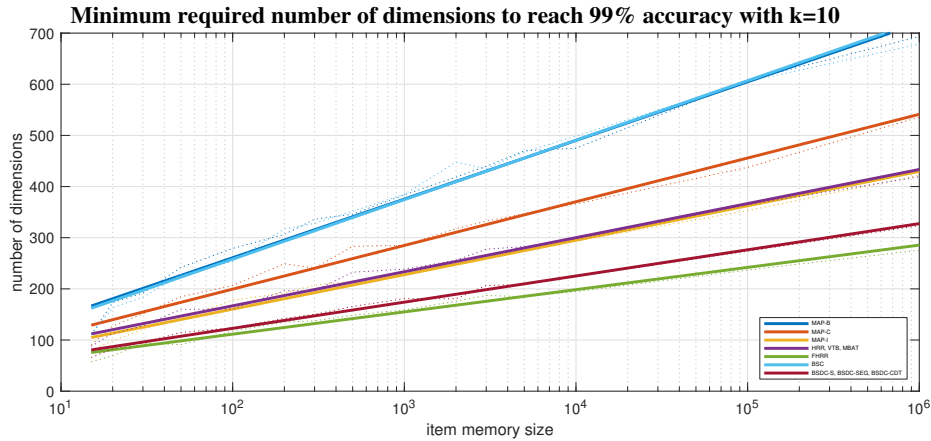}
\caption{Result of the capacity experiment with fixed number of neighbors and varying item memory size. Please note the logarithmic scale. The straight lines are fitted exponential functions.}
\label{img:bundle_results_item_sizes}
\end{figure}

The results can be seen in the Fig.~\ref{img:bundle_results_item_sizes} (using a logarithmic scale for the item memory size). 
Although the absolute performance varies between VSAs, the shape of the curves are in accordance with Plate's previous experiment on HRRs.
Since there are no qualitative differences between the VSAs (the ordering of the graphs is consistent), our above comparison of VSAs for a varying number of bundled vectors $k$ is presumably representative also for other item memory sizes $N$.

\subsection{Performance of approximately invertible binding} 
\label{subsec:appr_unbind}

The taxonomy in Fig.~\ref{fig: taxonomy} includes three VSAs that only have an approximate inverse binding: \textbf{MAP-C}, \textbf{VTB} and \textbf{HRR}. 
The question is: \textit{How good is the approximation of the binding inverse?}
To evaluate the performance of the approximate inverses, we use a setup similar to \cite{Gosmann2019}. 
We extended the experiment to compare the accuracy of approximate unbinding of the three relevant VSAs. 
The experiment is defined as follows: we start with an initial random vector \textbf{v} and bind it sequentially with $n$ other random vectors $\mathbf{r_1 \cdots r_n}$ to an encoded sequence \textbf{S} (see eq.~\ref{eq:seq_bind}).
The task is to retrieve the elemental vector \textbf{v} by sequentially unbinding the random vectors $\mathbf{r_1 \cdots r_n}$ from \textbf{S}.
The result is a vector $\mathbf{v^{\prime}}$ that should be highly similar to the original vector \textbf{v} (see eq.~\ref{eq:seq_unbinding}). 

\begin{equation}
\mathbf{S}=((\mathbf{v} \otimes \mathbf{r_1}) \otimes \mathbf{r_2})... \otimes \mathbf{r_n}
\label{eq:seq_bind}
\end{equation}

\begin{equation}
\mathbf{v^{\prime}}=\mathbf{r_1} \oslash ... (\mathbf{r_{n-1}} \oslash (\mathbf{r_n} \oslash \mathbf{S}))
\label{eq:seq_unbinding}
\end{equation}

\begin{figure}
\centerline{\includegraphics[width=\textwidth]{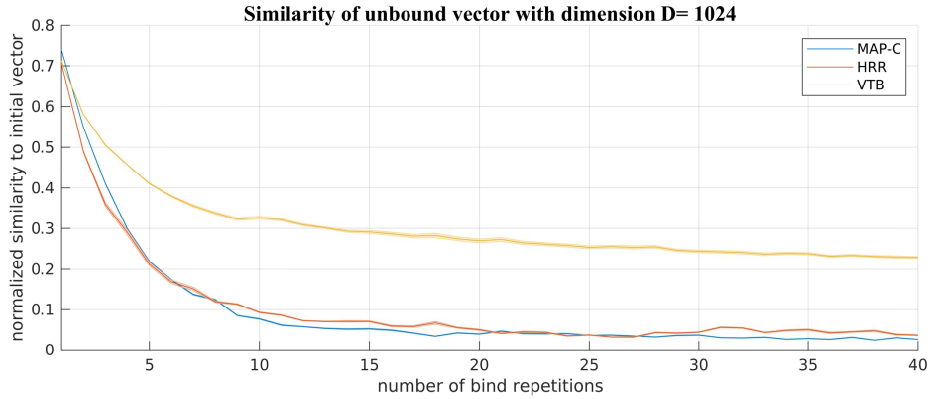}}
\caption{Normalized similarity between the initial vector \textbf{v} and the unbound sequence vector $\mathbf{v^{\prime}}$ with different numbers of sequences.} 
\label{img:bind_unbind_results}
\end{figure}

We applied the described procedure for the 3 approximated VSAs (all exact-invertible bindings would produce 100\% accuracy and are not shown in the plots) with $n=40$ sequences and $D=1024$ dimensions. 
The evaluation criterion is the similarity of $v$ and $v^{\prime}$, normalized to range $[0,1]$ (minimum to maximum possible similarity value).
Results are shown in Fig.~\ref{img:bind_unbind_results}.
In accordance with the results from \cite{Gosmann2019}, the \textbf{VTB} binding and unbinding performs better than the circular convolution/correlation from \textbf{HRR}. 
It reaches the highest similarity over the whole range.
The bind/unbind operator of the \textbf{MAP-C} architecture with values within the range $[-1,1]$ performs slightly worse than \textbf{HRR}. 
In practice, VSA systems with such long sequences of approximate unbindings can incorporate a denoising mechanism. For example, a nearest neighbor search in an item memory with atomic vectors to clean up the resulting vector (often referred to as clean-up memory). 
 
\subsection{Unbinding of bundled pairs} 

The third experiment combines the bundling, the binding and the unbinding operator in one scenario. 
It extends the example from the introduction, where we bundled three role-filler pairs to encode the knowledge about one country. A VSA allows querying for a filler by unbinding the role. 
Now, the question is: \textit{How many property-value (role-filler) pairs can be bundled and still provide the correct answer to any query  by unbinding a role?}
This is similar to unbinding of a noisy representation and to the experiment on scaling properties of VSAs in \cite[p.~141]{Eliasmith2013} but using only a single item memory size. 

Similar to the bundle capacity experiment in the previous section \ref{subsec:bundle_cap}, we create a database (item memory) of $N=1,000$ random elemental vectors.
We combine $2 k$ ($k$ roles and $k$ fillers) randomly chosen elementary vectors from the item memory to $k$ vector pairs by binding these two entities.
The result are $k$ bound pairs, equivalent to the property-value pairs from the USA example ($Name \otimes USA$...).
These pairs are bundled to a single representation $R$ (analog to the representation $R_{USA}$)
which creates a noisy version of all bound pairs. 
The goal is to retrieve all $2  k$ elemental vectors from the compact hypervector $R$ by unbinding. 
The evaluation criterion is defined as follows: we compute the ratio (accuracy) of correctly recovered vectors to the number of all initial vectors ($2 k$).
As in the capacity experiment, we used a variable number of dimensions $D=4 ... 1156$ and a varying number of bundled pairs $k=2 ... 50$. 
Finally, we run the experiment 10 times and use the mean values.

\begin{figure}
\centerline{\includegraphics[width=\textwidth]{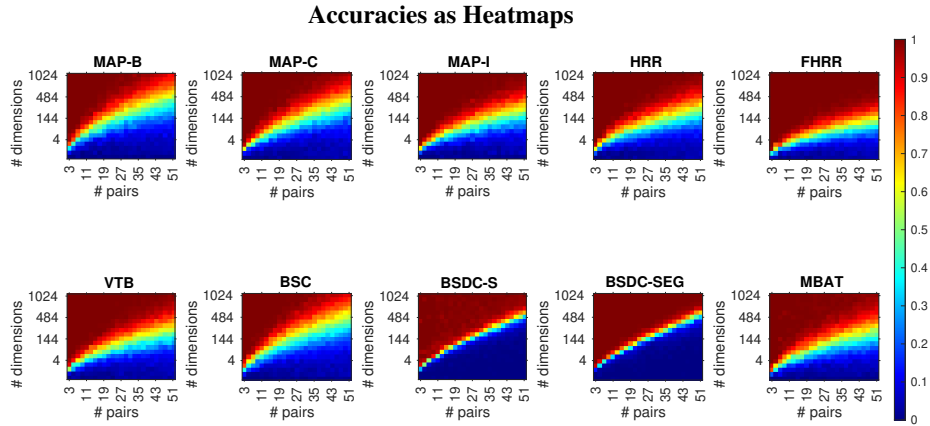}}
\caption{Heat-maps showing the accuracies of different number of bundled vectors and numbers of dimensions.} 
\label{img:bind_results_2d}
\end{figure}

\begin{figure}
\centerline{\includegraphics[width=\textwidth]{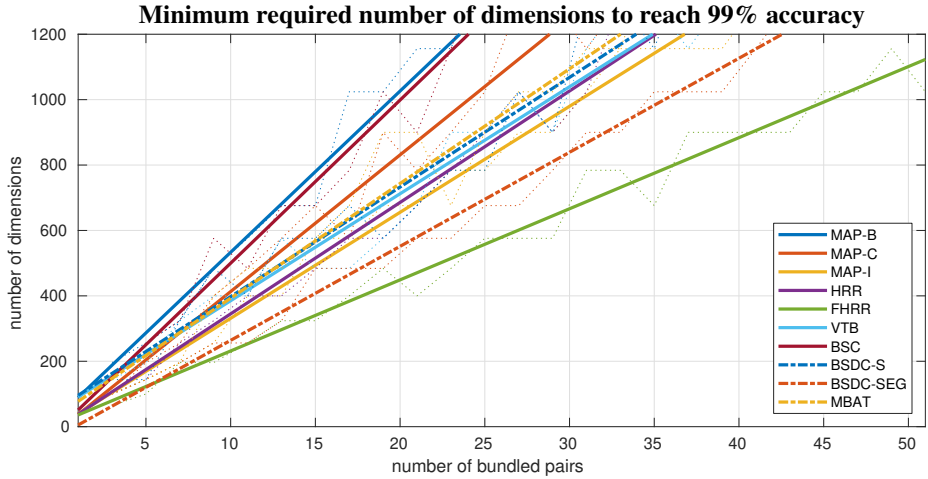}}
\caption{Minimum required number of dimensions to reach 99\% accuracy in unbinding of bundled pairs experiment. The solid lines represent linear fitted curves.} 
\label{img:bind_results}
\end{figure}

Similar to the bundling capacity experiment (Sec.~\ref{subsec:bundle_cap}), we provide two plots: Fig.~\ref{img:bind_results_2d} presents the accuracies as heat-maps for all combinations of numbers of bundled pairs and dimensions, and Fig.~\ref{img:bind_results} shows the minimum required number of dimensions to achieve 99\% accuracy. 
Interestingly, the overall appearance of the heatmaps of the two BSDC architectures in Fig.~\ref{img:bind_results_2d} is roughly the same, but the \textbf{BSDC-SHIFT} has a noisy red area, which means that some retrievals failed even if the number of dimensions is high enough in general. 
The similar fuzziness can be seen at the heat-map of the \textbf{MBAT} VSA.  

Again, Fig.~\ref{img:bind_results} summarizes the results to 1-D curves. It contains more curves than in the previous section because some VSAs share the same bundling operator, but each has an individual binding operator. 
For example, the performance of the different \textbf{BSDC} architectures varies.
The sparse VSA with the segmental binding is more dimension-efficient than shifting the whole vector.
However, all BSDC variants are less dimension-efficient than \textbf{FHRR} in this experiment, although they performed similar in the capacity experiment from Fig.~\ref{img:bundle_results}.
Furthermore, all VSAs based on the normal (Gaussian) distributed continuous space (\textbf{HRR}, \textbf{VTB} and \textbf{MBAT}) achieve very similar results. 
It seems that matrix binding (e.g. \textbf{MBAT} and \textbf{VTB}) does not significantly improve the binding and unbinding. 

Finally, we evaluate the VSAs by comparing their accuracies to those of the capacity experiment from Sec.~\ref{subsec:bundle_cap} as follows:
We select the minimum required number of dimensions to retrieve either 15 bundled vectors (capacity experiment in sec.~\ref{subsec:bundle_cap}) or 15 bundled pairs (bound vectors experiment). 
Table~\ref{tab:min_dim_comparison} summarizes the results and shows the increase between the bundle and the binding-plus-bundle experiment. 
Noticeably, there is a significant rise of the number of dimensions for the sparse binary VSA. 
It requires up to 44\% larger vectors when using the bundling in combination with binding. 
However, the segmental shifting method with an increase of 22\% works better than shifting the whole vector. 
One reason could be the increasing density during binding of sparsely distributed vectors because it uses only the disjunction without a thinning procedure. 
\textbf{MAP-C}, \textbf{MAP-B}, \textbf{MAP-I}, \textbf{HRR}, \textbf{FHRR} and \textbf{BSC} only show a marginal change of the required number of dimensions.
Again, the complex \textbf{FHRR} VSA achieves the overall best performance regarding minimum number of dimensions and increase in order to account for pairs.  
However, this might result mainly from the good bundling performance rather than the better binding performance. 

\begin{table}
\caption{Comparison of the minumum required number of dimensions to reach a perfect retrieval of 15 bundled vectors and 15 bundled pairs (results are rounded to the tenth unit). 4th column shows the growth between the first and the second experimental results (rounded to one unit).}\label{tab:min_dim_comparison}
\begin{center}
\begin{tabular}{p{1.8cm}|p{2.2cm}|p{2cm}|r}
\hline
\textbf{Vector space} &  \textbf{\# Dimensions to bundle 15 vectors} & \textbf{\# Dimensions to bundle 15 pairs} & \textbf{Increase}\\
\specialrule{1.5pt}{1pt}{1pt}
MAP-C & 640 & 620 & -3\%\\
MAP-B & 790 & 780 & -1\%\\
BSC & 750 & 750 & $\pm 0$\%\\ 
HRR & 510 & 520 & +2\%\\
FHRR & 330 & 340 & +3\%\\
MAP-I & 470 & 490 & +4\% \\
VTB  & 510 & 550 & +7\%\\
MBAT & 510 & 570 & +11\%\\
BSDC-SEG & 320 & 410 & +22\%\\
BSDC-S & 320 & 570 & +44\%\\
\hline
\end{tabular}
\end{center}
\end{table}

\section{Practical Applications}
\label{sec:apps}

This section experimentally evaluates the different VSAs on two practical applications. The first is recognition of the language of a written text. The second is a task from mobile robotics: visual place recognition using real-world images, e.g., imagery of a 2800 km journey through Norway across different seasons.
We chose these practical applications since the former is an established example from the VSA literature and the latter an example of a combination of VSAs with Deep Neural Networks.
Again, we will compare VSA using the same number of dimensions. The actual memory consumption and computational cost per dimension can be quite different for each VSA. However, this will strongly depend on the available hard- and software.

\subsection{Language Recognition}
\label{subsec:lang_rec}
For the first application, we selected a task that has previously been addressed using a VSA in the literature: recognizing the language of a written text. 
For instance, \cite{Joshi2017} presents a VSA approach to recognize the language of a given text from 21 possible languages.
Each letter is represented by a randomly chosen hypervector (a vector symbolic representation). 
To construct a meaningful representation of the whole language, short sequences of letters are combined in \textit{n-grams}. 
The basic idea is to use VSA operations (binding, permutation, and bundling) to create the n-grams and compute an item memory vector for each language.
The used permutation operator $\rho$ is a simple shifting of the whole vector by a particular amount (e.g., permutation of order 5 is written as $\rho^5$).
For example, the encoding of the word 'the' in a 3-gram (that combine exactly the three consecutive letters) is done as follows:

\begin{enumerate}
\item Basis is a fixed random hypervector for each letter: $\mathbf{v_t}, \ \mathbf{v_h}, \ \mathbf{v_e}$
\item The vector of each letter in the n-gram is permuted with the permutation operator according to the position in the n-gram: $\rho^0 \mathbf{v_t}, \rho^1 \mathbf{v_h}, \rho^2 \mathbf{v_e}$ 
\item Permuted letter vectors are bound together to achieve a single vector that encodes the whole n-gram:\\ $\mathbf{V_{the}}= \rho^0 \mathbf{v_t} \otimes \rho^1 \mathbf{v_h} \otimes  \rho^2 \mathbf{v_e} $ 
\end{enumerate}

The ``learning'' of a language is simply done by bundling all n-grams of a training dataset ($\mathbf{V_{english}}=\mathbf{V_{the}} + \mathbf{V_{...}} $). 
The result is a single vector representing the n-gram statistics of this language (i.e., the multiset of n-grams) and that can then be stored in an item memory.
To later recognize the language of a given query text, the same procedure as for learning a language is repeated to obtain a single vector that represents all n-grams in the text, and a nearest neighbor query with all known language vectors in the item memory is performed. 

We use the experimental setup from \cite{Joshi2017} with 21 languages and 3-grams to compare the performance of the different available VSAs.
Since the matrix binding VSAs need a lot of time to learn the whole language vectors with our current implementation, we used a fraction of 1,000 training and 100 test sentences per language (which is 10\% of the total dataset size from \cite{Joshi2017}).

\begin{figure}
\centerline{\includegraphics[width=\textwidth]{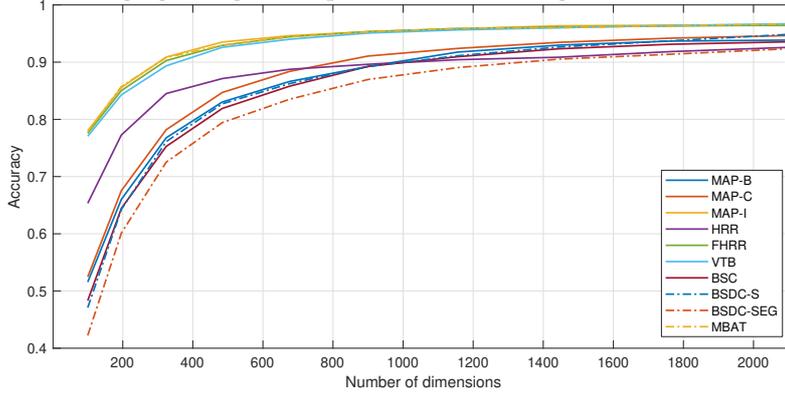}}
\caption{Accuracy on the language recognition experiment with increasing number of dimensions. The results are smoothed with an average filtering with kernel size of three. } 
\label{img:lang_rec_results}
\end{figure}

Fig.~\ref{img:lang_rec_results} shows the achieved accuracy of the different VSAs at the language recognition task for a varying number of dimensions between 100 and 2,000.
In general, the more dimensions are used, the higher is the achieved accuracy.
\textbf{MBAT}, \textbf{VTB} and \textbf{FHRR} need fewer dimensions to achieve high accuracy. 
It can be seen that the \textbf{VTB} binding is considerably better at this particular task than the original circular convolution binding of the \textbf{HRR} architecture (\textbf{HRR} is less efficient compared to \textbf{VTB}).
Interestingly, the \textbf{FHRR} has almost the same accuracy as the architectures with matrix binding (\textbf{VTB} and \textbf{MBAT}) although it uses less costly element-wise operations for binding and bundling. 
Finally, \textbf{BSDC-CDT} was not evaluated on this task. Since it has no thinning process after bundling, bundling hundreds of n-gram vectors results in an almost completely filled vector which is unsuited for this task.

\subsection{Place Recognition}
\label{subsec:place_recognition}

Visual place recognition is an important problem in the field of mobile robotics, e.g., it is an important means for loop closure detection in SLAM (Simulation Localization And Mapping).
The following Sec.~\ref{par:comp_db_query} will introduce this problem and outline the state-of-the-art approach SeqSLAM \cite{milford12}.
In \cite{Neubert2019}, we already described how a VSA can be used to encode the information from a sequence of images in a single hypervector and perform place recognition similarly to SeqSLAM. 
Approaching this problem with a VSA is particularly promising since the image comparison is typically done based on the similarity of high-dimensional image descriptor vectors.
The VSA approach has the advantage of only requiring a single vector comparison to decide about a matching -- while SeqSLAM typically requires 5-10 times as many comparisons.
After presentation of the CNN-based image encodings in Sec.~\ref{par:image_encoding}, Sec.~\ref{par:sequence_vsa} will use this procedure from \cite{Neubert2019} to evaluate the performance of the different VSAs.

\subsubsection{Pairwise descriptor comparison and SeqSLAM} 
\label{par:comp_db_query} 
\begin{figure}[t]
\centerline{\includegraphics[width=\textwidth]{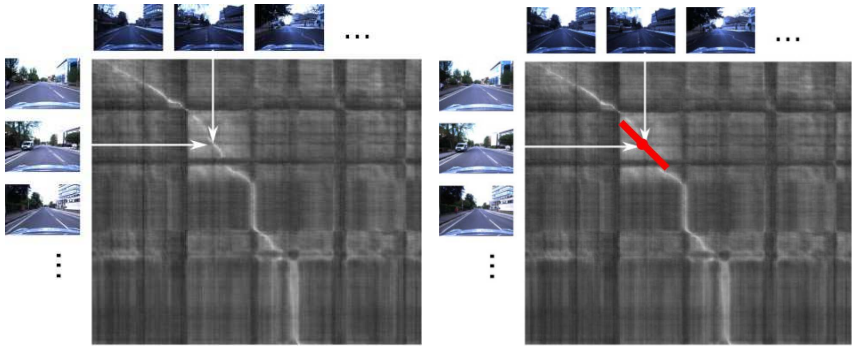}}
\vspace{-0.3cm}
\caption{Evaluation metric of the place recognition experiment. The gray tones images represent the similarity matrix (color encoded similarities between the database and query images -- bright pixels corresponding to a high similarity). Left: pairwise comparison of database and query images. Right: sequence-based comparison of query and database images with the red line representing the sequence of compared images.} 
\label{img:pr_vis}
\end{figure}

Place recognition is the problem of associating the robot's current camera view with one or multiple places from a database of images of known places (e.g., images of all previously visited locations).
The essential source of information is a descriptor for each image that can be used to compute the similarity between each pair of a database and a query image.
The result is a pairwise similarity matrix as illustrated on the left side of Fig.~\ref{img:pr_vis}.
The most similar pairs can then be treated as place matchings. 

Place recognition is a special case of image retrieval.
It differs from a general image retrieval task since the images typically have a temporal and spatial ordering -- we can expect temporally neighbored images to show spatially neighbored places.
A state-of-the-art place recognition method that exploits this additional constraint is SeqSLAM \cite{milford12}, which evaluates short sequences of images in order to find correspondences between the query camera stream and the database images.
Basically, SeqSLAM not only compares the current camera image to the database, but also the previous (and potentially the subsequent) images.

\begin{algorithm}[H] 
 \footnotesize 
 \begin{algorithmic}[1]
 \renewcommand{\algorithmicrequire}{\textbf{Input:}}
 \renewcommand{\algorithmicensure}{\textbf{Output:}}
 \REQUIRE Similarity matrix $S$ of size $m\times n$, sequence length parameter $d$
 \ENSURE  New similarity matrix $R$  
  \FOR{$i = 1:m$}
    \FOR{$j = 1:n$}    
      \STATE{accSim = 0}    
      \FOR{$k = -d:1:d$}
	\STATE{ accSim += S(i+k, j+k) }
      \ENDFOR      
      \STATE{ R(i,j) = accSim / (2$\cdot$d+1) }      
    \ENDFOR  
  \ENDFOR
  \RETURN{$R$}
 \end{algorithmic}
 \caption{Simplified SeqSLAM core algorithm} 
 \label{alg:SeqSLAM}
\end{algorithm}

Algorithm~\ref{alg:SeqSLAM} illustrates the core processing of SeqSLAM in a simplified algorithmic listing.
Input is a pairwise similarity matrix $S$.
In order to exploit the sequential information, the algorithm iterates over all entries of $S$ (the loops in lines 1 and 2). 
For each element the average similarities over the sequence of neighbored elements is computed in a third loop (line 4).
This neighborhood sequence is illustrated as a red line in Fig.~\ref{img:pr_vis} (basically, this is a sparse convolution).
This simple averaging is known to significantly improve the place recognition performance, in particular in case of changing environmental conditions \cite{milford12}.
The listing is intended to illustrate the core idea of SeqSLAM. It is simplified since border effects are ignored and since the original SeqSLAM evaluates different possible velocities (i.e. slopes of the neighborhood sequences). For more details, please refer to \cite{milford12}.
The key benefit of the VSA approach to SeqSLAM is that it will allow to completely remove the inner-loop.

\subsubsection{Evaluation procedure} 
\label{par:eval} 

To compare the performance of different place recognition approaches in our experiments, we use a standard evaluation procedure based on ground-truth information about place matchings \cite{Neubert2019a}.
It is based on five datasets with available ground truth:
\textbf{StLucia} Various Times of the Day \cite{stlucia}, \textbf{Oxford} RobotCar \cite{robotcar}, \textbf{CMU} Visual Localization \cite{cmu}, \textbf{Nordland} \cite{nordland} and \textbf{Gardens} Point Walking \cite{gardenspoint}.
Given the output of a place recognition approach on a dataset (i.e., the initial matrix of pairwise similarities $S$ or the output of SeqSLAM $R$), we run a series of thresholds on the similarities to get a set of binary matching decisions for each individual threshold. 
We use the ground truth to count true-positive (TP), false-positive (FP), and false-negative (FN) matchings, and further compute a point on the precision-recall curve for each threshold with precision $P=TP/(TP+FP)$ and recall $R=TP/(TP+FN)$.
To obtain a single number that represents the place recognition performance, we report AUC, the area under the precision-recall curve (i.e., average precision, obtained by trapezoidal integration). 

\subsubsection{Encoding Images for VSAs}
\label{par:image_encoding}

Using VSAs in combination with real-world images for place recognition requires an image encoding into meaningful descriptors. 
Dependent on the particular vector space of the VSA the encoding will be different.
We will first describe the underlying basic image descriptor, followed by an explanation and evaluation of the individual encodings for each VSA.

\begin{figure}
\centerline{\includegraphics[width=\textwidth]{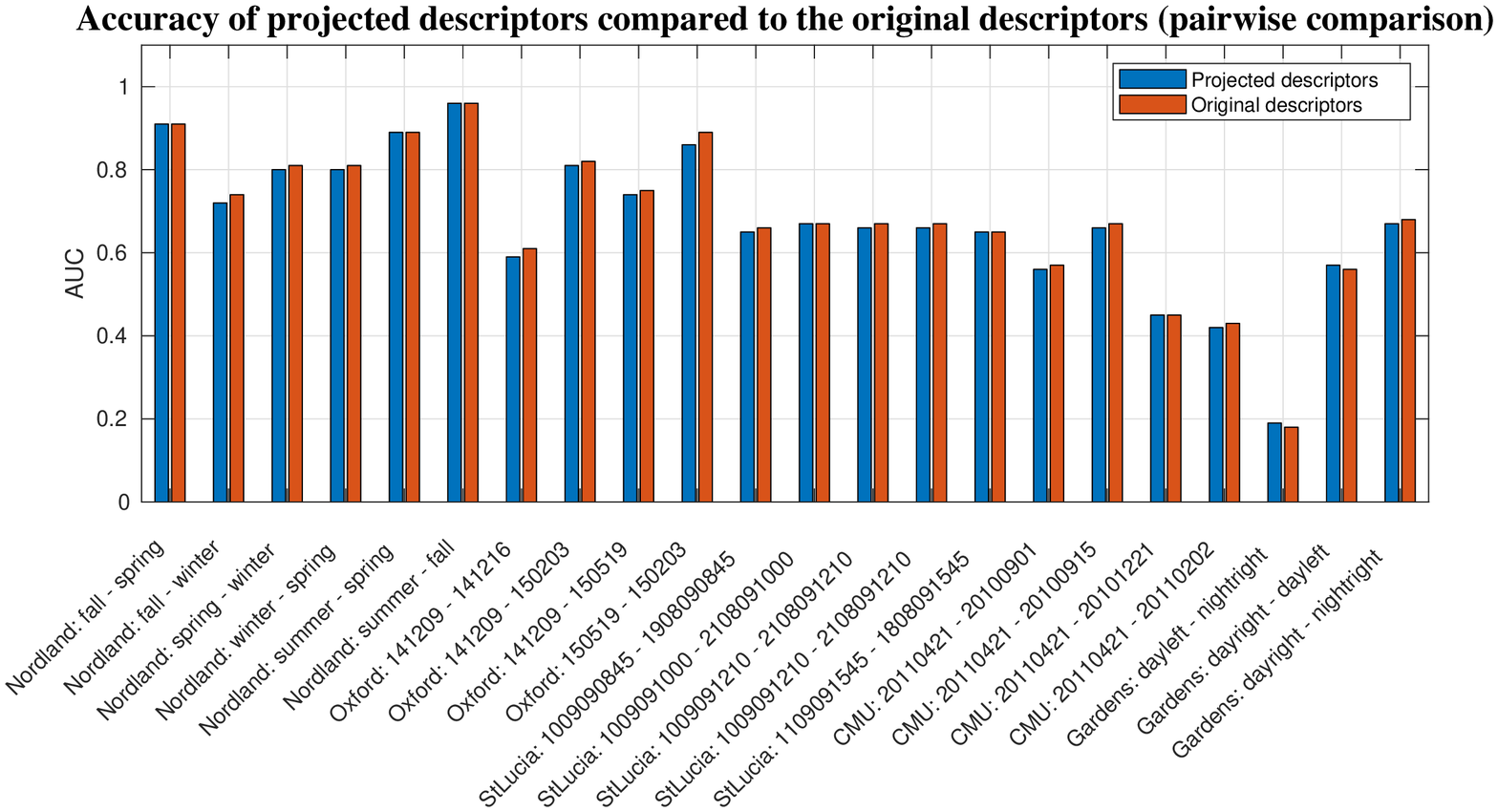}}
\caption{AUC of original descriptors and projected descriptors (decrease number of dimensions) for each dataset. Evaluation based on pairwise comparison of database and query images.} 
\label{img:effect_dim_reduction}
\end{figure}

We use a basic descriptor similar to our previous work \cite{Neubert2019a}.
S\"underhauf et al. \cite{Sunderhauf2015} showed that early convolutional layers of CNNs are a valuable source for creating robust image descriptors for place recognition. 
For example, the pre-trained AlexNet \cite{Krizhevsky12} generates the most robust image descriptors at the third convolution level. 
To use these as input for the place recognition pipeline, all images pass through the first three layers of AlexNet and the output tensor of size of $13\times13\times384$ is flattened to a vector of size 64,896. 
Next, we apply a dimension-wise standardization of the descriptors for each dataset following Schubert et al. \cite{Schubert2020}.
Although this is already a high-dimensional vector, we use random projections in order to distribute information across dimensions and influence the number of dimensions:
To obtain a $N$-dimensional vector (e.g. $N=4,096$) from a $M$-dimensional space (e.g. $M=64,896$), the original vector is multiplied by a random $M \times N$ matrix with values drawn from a Gaussian normal distribution. $M$ is row-wise normalized. 
Such a dimensional reduction can lead to loss of information. 
The effect on the pairwise place recognition performance for each data set is shown in Fig.~\ref{img:effect_dim_reduction}. 
It shows the AUC of pairwise comparison of both, the original descriptors and the dimension-reduced descriptors (calculated and evaluated as described in the section above). 
The plot supports that the random projection is a suitable method to reduce the dimensionality and distribute information, since the projected descriptors reach almost the same AUC as the original descriptors.

Afterwards the descriptors can be converted into the vector spaces of the individual VSAs (cf. table~\ref{tab:VSAs}).
Table~\ref{tab:encodings} lists the encoding methods to convert the projected, standardized CNN descriptors to the different VSA vector spaces.
It has to be noticed that the sLSBH method doubled the number of dimensions of the input vector (pleaser refer to \cite{Neubert2019a} for details).
The table also lists the influence of the encodings on the place recognition performance (mean and standard deviation of AUC change over all datasets).
The performance change in the 4th column was computed by $(Acc_{projected}-Acc_{converted}) / Acc_{projected}$.

It can be seen that the encoding method for \textbf{HRR}, \textbf{VTB} and \textbf{MBAT} VSAs does not influence the performance.
In contrast, the conversion of the real-valued space into the sparse binary domain leads to significant performance losses (approx. 22\%). 
However, this is mainly due to the fact that we compare the encoding of a \textit{dense real} valued vector into a \textit{sparse binary} vector of only twice the number of dimensions (a property of the used sLSBH procedure \cite{Neubert2019a}). The encoding quality improves, if the number of dimensions in the sparse binary vector is increased. However, for consistency reasons, we keep the number of dimensions fixed. 
The density of the resulting sparse vectors is $1/\sqrt{2\cdot D}$.

\begin{table}
\begin{center} 
\caption{Encoding methods. Last column represents the AUC change between the original data (after projection) and the converted data with pairwise comparison. The density of sLSBH is $1/\sqrt{2 \cdot D}$.} 
\label{tab:encodings}
\begin{tabular}{p{3cm}|p{3cm}|p{2.5cm}|C{2.25cm}}
\hline
\rule{0pt}{10pt}   
\multirow{2}{*}{\textbf{elements $X$ of}  \textbf{Space $V$}} & \multirow{2}{*}{\textbf{Encoding of input $I$}} & \multirow{2}{*}{\textbf{VSA}} & \multirow{2}{*}{\textbf{perf.}  \textbf{change [\%]}} \\
&&& \\
\specialrule{1.5pt}{1pt}{1pt}
$X \in\{-1,1\}^D$ & $X= \begin{cases} 1 & I>0 \\ -1 & I <= 0 \end{cases} $ & MAP-B & $-2.2 \pm1.9$ \\\hline
$X \in\{0,1\}^D$ & $X= \begin{cases} 1 & I>0 \\ 0 & I <= 0 \end{cases} $ & BSC & $-2.2 \pm 1.9$ \\\hline
$X \in [-1, 1]^D$ & $X= \begin{cases} 1 & I>=1 \\ -1 & I <= -1 \\ I & else \end{cases} $ & MAP-C & $-1 \pm 1.3$\\\hline
\rule{0pt}{10pt}   
\multirow{2}{*}{$X \in \mathbb{R}^D$} & \multirow{2}{*}{$X=\frac{I}{norm(I)}$} & \multirow{2}{*}{\shortstack[l]{HRR, VTB, \\ MBAT}} & \multirow{2}{*}{0} \\
& & \\\hline
\rule{0pt}{10pt}   
\multirow{2}{*}{$X \in \mathbb{C}^D,$  $X=e^{i\cdot \theta}$} & \multirow{2}{*}{$\theta = arg(\mathcal{F}\{I\})$}& \multirow{2}{*}{FHRR} & \multirow{2}{*}{$-0.9 \pm 0.8$} \\
\rule{0pt}{10pt}   
&&\\\hline
\rule{0pt}{10pt}   
\multirow{2}{*}{$X \in \{0,1\}^{2 \cdot D} $} & \multirow{2}{*}{sLSBH \cite{Neubert2019}} & \multirow{2}{*}{BSDC-S, BSDC-SEG} & \multirow{2}{*}{$-21.9\pm16$} \\
&&\\
\hline
\end{tabular}
\end{center}
\end{table}

\subsubsection{VSA SeqSLAM}
\label{par:sequence_vsa}

The key idea of the VSA implementation of SeqSLAM is to replace the costly post-processing of the similarity matrix $S$ in Algorithm~\ref{alg:SeqSLAM} by a superposition of the information of neighbored images already in the high-dimensional descriptor vector of an image.
Thus, the sequential information can be harnessed in a simple pairwise descriptor comparison and the inner-loop of SeqSLAM (line 4 in Algorithm~\ref{alg:SeqSLAM}) becomes obsolete.

This idea can be implemented as preprocessing of descriptors before the computation of the pairwise similarity matrix $S$.
Each descriptor $X_i$ in the database and query set is processed independently into an new descriptor vector $Y_i$ that also encodes the neighboring descriptors:
\begin{equation}
Y_{i}=+_{k=-d}^{d}\left(X_{i+k} \otimes P_{k}\right)
\label{eq:VSA_sequence}
\end{equation}

Each image descriptor from the sequence neighborhood is bound to a static position vector $P_k$ before bundling to encode the ordering of the images within the sequence.
The position vectors are randomly chosen, but fixed across all database and query images.
In a later pairwise comparison of two such vectors $Y$, only those descriptors $X$ that are at corresponding positions within the sequence contribute to the overall similarity (due to the quasi-orthogonality of the random position vectors and the properties of the binding operator).
In the following, we will evaluate the place recognition performance when implementing this approach with the different VSAs. Please refer to \cite{Neubert2019a} for more details on the approach itself.

\subsubsection{Results}

In the experiments, we use 4,096 dimensional vectors (except for sLSBH encodings with twice this number) and sequence length $d=5$. 
Table~\ref{tab:pr_results} shows the results when using either the original SeqSLAM on an particular encoding or the VSA-implementation.
The performance of the original SeqSLAM on the original descriptors (but with dimensionality reduction and standardization) can, e.g. be seen at the VTB column.
To increase the readability, we highlighted the overall best results in bold and visualized the relative performance of a VSA to the corresponding original SeqSLAM with colored arrows. 
In most cases, the VSA approaches can approximate the SeqSLAM method with essentially the same AUC. 
Particularly the real-valued vector spaces (\textbf{MAP-C}, \textbf{HRR}, \textbf{VTB}) yield good AUC in both the encoding itself (Table~\ref{tab:encodings}) and the sequence-based place recognition task. 
\textbf{MAP-C} achieves 100\% AUC on the Nordland dataset (which is even slightly better than the SeqSLAM algorithm) and has no considerable AUC reduction in any other datasets.
Also the \textbf{VTB} and \textbf{MBAT} architectures achieve very similar results to the original SeqSLAM approach. 
However, it has to be noticed that these VSAs use matrix binding methods, which leads to a high computational effort compared to element-wise binding operations. 
The performance of the sparse VSAs (\textbf{BSDC-S}, \textbf{BSDC-SEG}) varies, including cases where the performance is considerably worse than the original SeqSLAM (which in turn achieves surprisingly good results given the overall performance drop of the sparse encoding from Table~\ref{tab:encodings}).

\begin{landscape}
\begin{table}[t]
\caption{Results (AUC) of the Place Recognition Experiment with original datasets. Sequence length is 5 and all sequence methods use constant velocity. The colored arrows indicate large ($\ge$25\%), medium ($\ge$10\%), or no ($<$10\%) deviation from SeqSLAM (orig.).}
\tiny
\begin{tabular}{p{0.55cm}|p{0.8cm}|p{0.8cm}||p{0.15cm}p{0.45cm}|p{0.15cm}p{0.45cm}|p{0.15cm}p{0.45cm}|p{0.15cm}p{0.45cm}|p{0.15cm}p{0.45cm}|p{0.15cm}p{0.45cm}|p{0.15cm}p{0.45cm}|p{0.15cm}p{0.45cm}|p{0.15cm}p{0.45cm}|p{0.15cm}p{0.45cm}}
\textbf{Dataset} & \textbf{Database} & \textbf{Query} & \multicolumn{2}{c}{\textbf{MAP-B}} & \multicolumn{2}{c}{\textbf{MAP-C}} & \multicolumn{2}{c}{\textbf{MAP-I}} & \multicolumn{2}{c}{\textbf{HRR}} & \multicolumn{2}{c}{\textbf{VTB}} & \multicolumn{2}{c}{\textbf{MBAT}} & \multicolumn{2}{c}{\textbf{FHRR}} & \multicolumn{2}{c}{\textbf{BSC}} & \multicolumn{2}{c}{\textbf{BSDC-S}} & \multicolumn{2}{c}{\textbf{BSDC-SEG}} \\ 
  &   &   & orig & VSA & orig & VSA & orig & VSA & orig & VSA & orig & VSA & orig & VSA & orig & VSA & orig & VSA & orig & VSA & orig & VSA \\ 
\hline 
Nordland & fall & spring & 0.98~ & \textbf{1.00}~ \color{equal}{$\rightarrow$}  & 0.98~ & \textbf{1.00}~ \color{equal}{$\rightarrow$}  & 0.98~ & \textbf{1.00}~ \color{equal}{$\rightarrow$}  & 0.98~ & \textbf{1.00}~ \color{equal}{$\rightarrow$}  & 0.98~ & \textbf{1.00}~ \color{equal}{$\rightarrow$}  & 0.98~ & \textbf{1.00}~ \color{equal}{$\rightarrow$}  & 0.98~ & \textbf{1.00}~ \color{equal}{$\rightarrow$}  & 0.98~ & \textbf{1.00}~ \color{equal}{$\rightarrow$}  & 0.92~ & 0.97~ \color{equal}{$\rightarrow$}  & 0.92~ & 0.97~ \color{equal}{$\rightarrow$}  \\ 
Nordland & fall & winter & 0.98~ & 0.99~ \color{equal}{$\rightarrow$}  & 0.99~ & \textbf{1.00}~ \color{equal}{$\rightarrow$}  & 0.98~ & 0.99~ \color{equal}{$\rightarrow$}  & 0.99~ & \textbf{1.00}~ \color{equal}{$\rightarrow$}  & 0.99~ & 0.99~ \color{equal}{$\rightarrow$}  & 0.99~ & \textbf{1.00}~ \color{equal}{$\rightarrow$}  & 0.98~ & \textbf{1.00}~ \color{equal}{$\rightarrow$}  & 0.98~ & 0.98~ \color{equal}{$\rightarrow$}  & 0.89~ & 0.87~ \color{equal}{$\rightarrow$}  & 0.89~ & 0.85~ \color{equal}{$\rightarrow$}  \\ 
Nordland & spring & winter & 0.96~ & 0.97~ \color{equal}{$\rightarrow$}  & 0.96~ & \textbf{0.99}~ \color{equal}{$\rightarrow$}  & 0.96~ & \textbf{0.99}~ \color{equal}{$\rightarrow$}  & 0.96~ & 0.98~ \color{equal}{$\rightarrow$}  & 0.96~ & 0.98~ \color{equal}{$\rightarrow$}  & 0.96~ & \textbf{0.99}~ \color{equal}{$\rightarrow$}  & 0.96~ & \textbf{0.99}~ \color{equal}{$\rightarrow$}  & 0.96~ & 0.97~ \color{equal}{$\rightarrow$}  & 0.82~ & 0.83~ \color{equal}{$\rightarrow$}  & 0.82~ & 0.84~ \color{equal}{$\rightarrow$}  \\ 
Nordland & winter & spring & 0.96~ & 0.97~ \color{equal}{$\rightarrow$}  & 0.96~ & \textbf{0.99}~ \color{equal}{$\rightarrow$}  & 0.96~ & \textbf{0.99}~ \color{equal}{$\rightarrow$}  & 0.96~ & 0.98~ \color{equal}{$\rightarrow$}  & 0.96~ & 0.98~ \color{equal}{$\rightarrow$}  & 0.96~ & \textbf{0.99}~ \color{equal}{$\rightarrow$}  & 0.96~ & \textbf{0.99}~ \color{equal}{$\rightarrow$}  & 0.96~ & 0.97~ \color{equal}{$\rightarrow$}  & 0.84~ & 0.83~ \color{equal}{$\rightarrow$}  & 0.94~ & 0.95~ \color{equal}{$\rightarrow$}  \\ 
Nordland & summer & spring & 0.98~ & 0.99~ \color{equal}{$\rightarrow$}  & 0.98~ & \textbf{1.00}~ \color{equal}{$\rightarrow$}  & 0.98~ & \textbf{1.00}~ \color{equal}{$\rightarrow$}  & 0.98~ & \textbf{1.00}~ \color{equal}{$\rightarrow$}  & 0.98~ & \textbf{1.00}~ \color{equal}{$\rightarrow$}  & 0.98~ & \textbf{1.00}~ \color{equal}{$\rightarrow$}  & 0.98~ & 0.99~ \color{equal}{$\rightarrow$}  & 0.98~ & \textbf{1.00}~ \color{equal}{$\rightarrow$}  & 0.93~ & 0.97~ \color{equal}{$\rightarrow$}  & 0.93~ & 0.97~ \color{equal}{$\rightarrow$}  \\ 
Nordland & summer & fall & \textbf{1.00}~ & \textbf{1.00}~ \color{equal}{$\rightarrow$}  & \textbf{1.00}~ & \textbf{1.00}~ \color{equal}{$\rightarrow$}  & \textbf{1.00}~ & \textbf{1.00}~ \color{equal}{$\rightarrow$}  & \textbf{1.00}~ & \textbf{1.00}~ \color{equal}{$\rightarrow$}  & \textbf{1.00}~ & \textbf{1.00}~ \color{equal}{$\rightarrow$}  & \textbf{1.00}~ & \textbf{1.00}~ \color{equal}{$\rightarrow$}  & \textbf{1.00}~ & \textbf{1.00}~ \color{equal}{$\rightarrow$}  & \textbf{1.00}~ & \textbf{1.00}~ \color{equal}{$\rightarrow$}  & 0.97~ & 0.99~ \color{equal}{$\rightarrow$}  & 0.97~ & 0.99~ \color{equal}{$\rightarrow$}  \\ 
\hline 
Oxford & 141209 & 141216 & \textbf{0.67}~ & 0.52~ \color{bad}{$\searrow$}  & 0.66~ & 0.64~ \color{equal}{$\rightarrow$}  & \textbf{0.67}~ & \textbf{0.67}~ \color{equal}{$\rightarrow$}  & 0.63~ & 0.53~ \color{bad}{$\searrow$}  & 0.63~ & 0.60~ \color{equal}{$\rightarrow$}  & 0.63~ & 0.61~ \color{equal}{$\rightarrow$}  & 0.63~ & 0.53~ \color{bad}{$\searrow$}  & \textbf{0.67}~ & 0.53~ \color{bad}{$\searrow$}  & 0.40~ & 0.36~ \color{bad}{$\searrow$}  & 0.40~ & 0.36~ \color{bad}{$\searrow$}  \\ 
Oxford & 141209 & 150203 & \textbf{0.85}~ & 0.79~ \color{equal}{$\rightarrow$}  & \textbf{0.85}~ & 0.82~ \color{equal}{$\rightarrow$}  & \textbf{0.85}~ & 0.81~ \color{equal}{$\rightarrow$}  & 0.84~ & \textbf{0.85}~ \color{equal}{$\rightarrow$}  & 0.84~ & 0.84~ \color{equal}{$\rightarrow$}  & 0.84~ & 0.80~ \color{equal}{$\rightarrow$}  & 0.84~ & 0.81~ \color{equal}{$\rightarrow$}  & \textbf{0.85}~ & 0.80~ \color{equal}{$\rightarrow$}  & 0.65~ & 0.55~ \color{bad}{$\searrow$}  & 0.65~ & 0.61~ \color{equal}{$\rightarrow$}  \\ 
Oxford & 141209 & 150519 & 0.82~ & 0.82~ \color{equal}{$\rightarrow$}  & 0.81~ & \textbf{0.83}~ \color{equal}{$\rightarrow$}  & 0.82~ & 0.75~ \color{equal}{$\rightarrow$}  & 0.80~ & 0.81~ \color{equal}{$\rightarrow$}  & 0.80~ & 0.76~ \color{equal}{$\rightarrow$}  & 0.80~ & 0.76~ \color{equal}{$\rightarrow$}  & 0.81~ & \textbf{0.83}~ \color{equal}{$\rightarrow$}  & 0.82~ & 0.80~ \color{equal}{$\rightarrow$}  & \textbf{0.83}~ & 0.81~ \color{equal}{$\rightarrow$}  & 0.71~ & 0.60~ \color{bad}{$\searrow$}  \\ 
Oxford & 150519 & 150203 & \textbf{0.91}~ & 0.89~ \color{equal}{$\rightarrow$}  & \textbf{0.91}~ & 0.90~ \color{equal}{$\rightarrow$}  & \textbf{0.91}~ & 0.88~ \color{equal}{$\rightarrow$}  & \textbf{0.91}~ & \textbf{0.91}~ \color{equal}{$\rightarrow$}  & \textbf{0.91}~ & 0.89~ \color{equal}{$\rightarrow$}  & \textbf{0.91}~ & \textbf{0.91}~ \color{equal}{$\rightarrow$}  & \textbf{0.91}~ & 0.90~ \color{equal}{$\rightarrow$}  & \textbf{0.91}~ & 0.89~ \color{equal}{$\rightarrow$}  & 0.84~ & 0.75~ \color{bad}{$\searrow$}  & 0.84~ & 0.67~ \color{bad}{$\searrow$}  \\ 
\hline 
StLucia & 100909\hbox{-}0845 & 190809\hbox{-}0845 & 0.83~ & 0.78~ \color{equal}{$\rightarrow$}  & 0.83~ & 0.83~ \color{equal}{$\rightarrow$}  & 0.83~ & 0.81~ \color{equal}{$\rightarrow$}  & 0.83~ & 0.83~ \color{equal}{$\rightarrow$}  & 0.83~ & 0.82~ \color{equal}{$\rightarrow$}  & 0.83~ & 0.82~ \color{equal}{$\rightarrow$}  & 0.83~ & 0.81~ \color{equal}{$\rightarrow$}  & 0.83~ & 0.79~ \color{equal}{$\rightarrow$}  & 0.83~ & 0.77~ \color{equal}{$\rightarrow$}  & \textbf{0.85}~ & 0.80~ \color{equal}{$\rightarrow$}  \\ 
StLucia & 100909\hbox{-}1000 & 210809\hbox{-}1000 & 0.87~ & 0.82~ \color{equal}{$\rightarrow$}  & \textbf{0.88}~ & 0.86~ \color{equal}{$\rightarrow$}  & 0.87~ & 0.85~ \color{equal}{$\rightarrow$}  & \textbf{0.88}~ & 0.86~ \color{equal}{$\rightarrow$}  & \textbf{0.88}~ & 0.86~ \color{equal}{$\rightarrow$}  & \textbf{0.88}~ & 0.86~ \color{equal}{$\rightarrow$}  & 0.87~ & 0.85~ \color{equal}{$\rightarrow$}  & 0.87~ & 0.82~ \color{equal}{$\rightarrow$}  & 0.86~ & 0.79~ \color{equal}{$\rightarrow$}  & 0.87~ & 0.82~ \color{equal}{$\rightarrow$}  \\ 
StLucia & 100909\hbox{-}1210 & 210809\hbox{-}1210 & \textbf{0.89}~ & 0.79~ \color{bad}{$\searrow$}  & 0.88~ & 0.84~ \color{equal}{$\rightarrow$}  & \textbf{0.89}~ & 0.83~ \color{equal}{$\rightarrow$}  & \textbf{0.89}~ & 0.84~ \color{equal}{$\rightarrow$}  & \textbf{0.89}~ & 0.84~ \color{equal}{$\rightarrow$}  & \textbf{0.89}~ & 0.84~ \color{equal}{$\rightarrow$}  & \textbf{0.89}~ & 0.84~ \color{equal}{$\rightarrow$}  & \textbf{0.89}~ & 0.80~ \color{bad}{$\searrow$}  & 0.88~ & 0.76~ \color{bad}{$\searrow$}  & 0.88~ & 0.77~ \color{bad}{$\searrow$}  \\ 
StLucia & 100909\hbox{-}1210 & 210809\hbox{-}1210 & \textbf{0.89}~ & 0.79~ \color{bad}{$\searrow$}  & 0.88~ & 0.84~ \color{equal}{$\rightarrow$}  & \textbf{0.89}~ & 0.83~ \color{equal}{$\rightarrow$}  & \textbf{0.89}~ & 0.84~ \color{equal}{$\rightarrow$}  & \textbf{0.89}~ & 0.84~ \color{equal}{$\rightarrow$}  & \textbf{0.89}~ & 0.84~ \color{equal}{$\rightarrow$}  & \textbf{0.89}~ & 0.84~ \color{equal}{$\rightarrow$}  & \textbf{0.89}~ & 0.80~ \color{bad}{$\searrow$}  & 0.88~ & 0.76~ \color{bad}{$\searrow$}  & 0.88~ & 0.77~ \color{bad}{$\searrow$}  \\ 
StLucia & 110909\hbox{-}1545 & 180809\hbox{-}1545 & \textbf{0.85}~ & 0.76~ \color{bad}{$\searrow$}  & \textbf{0.85}~ & 0.82~ \color{equal}{$\rightarrow$}  & \textbf{0.85}~ & 0.80~ \color{equal}{$\rightarrow$}  & \textbf{0.85}~ & 0.80~ \color{equal}{$\rightarrow$}  & \textbf{0.85}~ & 0.81~ \color{equal}{$\rightarrow$}  & \textbf{0.85}~ & 0.82~ \color{equal}{$\rightarrow$}  & \textbf{0.85}~ & 0.80~ \color{equal}{$\rightarrow$}  & \textbf{0.85}~ & 0.76~ \color{bad}{$\searrow$}  & 0.84~ & 0.75~ \color{bad}{$\searrow$}  & 0.84~ & 0.76~ \color{equal}{$\rightarrow$}  \\ 
\hline 
CMU & 20110421 & 20100901 & 0.64~ & \textbf{0.65}~ \color{equal}{$\rightarrow$}  & 0.63~ & 0.64~ \color{equal}{$\rightarrow$}  & 0.64~ & 0.60~ \color{equal}{$\rightarrow$}  & 0.63~ & 0.62~ \color{equal}{$\rightarrow$}  & 0.63~ & 0.61~ \color{equal}{$\rightarrow$}  & 0.63~ & 0.61~ \color{equal}{$\rightarrow$}  & 0.63~ & \textbf{0.65}~ \color{equal}{$\rightarrow$}  & 0.64~ & \textbf{0.65}~ \color{equal}{$\rightarrow$}  & 0.57~ & 0.50~ \color{bad}{$\searrow$}  & 0.57~ & 0.49~ \color{bad}{$\searrow$}  \\ 
CMU & 20110421 & 20100915 & \textbf{0.75}~ & 0.71~ \color{equal}{$\rightarrow$}  & 0.74~ & 0.74~ \color{equal}{$\rightarrow$}  & \textbf{0.75}~ & 0.70~ \color{equal}{$\rightarrow$}  & 0.74~ & 0.73~ \color{equal}{$\rightarrow$}  & 0.74~ & 0.73~ \color{equal}{$\rightarrow$}  & 0.74~ & 0.72~ \color{equal}{$\rightarrow$}  & 0.74~ & 0.73~ \color{equal}{$\rightarrow$}  & \textbf{0.75}~ & 0.72~ \color{equal}{$\rightarrow$}  & 0.66~ & 0.58~ \color{bad}{$\searrow$}  & 0.66~ & 0.56~ \color{bad}{$\searrow$}  \\ 
CMU & 20110421 & 20101221 & \textbf{0.56}~ & 0.54~ \color{equal}{$\rightarrow$}  & 0.55~ & 0.55~ \color{equal}{$\rightarrow$}  & \textbf{0.56}~ & 0.53~ \color{equal}{$\rightarrow$}  & 0.55~ & 0.54~ \color{equal}{$\rightarrow$}  & 0.55~ & 0.53~ \color{equal}{$\rightarrow$}  & 0.55~ & \textbf{0.56}~ \color{equal}{$\rightarrow$}  & \textbf{0.56}~ & \textbf{0.56}~ \color{equal}{$\rightarrow$}  & \textbf{0.56}~ & 0.55~ \color{equal}{$\rightarrow$}  & 0.31~ & 0.22~ \color{superbad}{$\downarrow$}  & 0.31~ & 0.20~ \color{superbad}{$\downarrow$}  \\ 
CMU & 20110421 & 20110202 & \textbf{0.52}~ & 0.45~ \color{bad}{$\searrow$}  & 0.51~ & 0.50~ \color{equal}{$\rightarrow$}  & \textbf{0.52}~ & 0.47~ \color{equal}{$\rightarrow$}  & 0.51~ & 0.48~ \color{equal}{$\rightarrow$}  & 0.51~ & 0.49~ \color{equal}{$\rightarrow$}  & 0.51~ & 0.48~ \color{equal}{$\rightarrow$}  & 0.51~ & 0.48~ \color{equal}{$\rightarrow$}  & \textbf{0.52}~ & 0.46~ \color{bad}{$\searrow$}  & 0.50~ & 0.37~ \color{superbad}{$\downarrow$}  & 0.50~ & 0.37~ \color{superbad}{$\downarrow$}  \\ 
\hline 
Gardens & day\hbox{-}left & night\hbox{-}right & 0.27~ & 0.19~ \color{superbad}{$\downarrow$}  & 0.26~ & 0.25~ \color{equal}{$\rightarrow$}  & 0.27~ & 0.22~ \color{bad}{$\searrow$}  & 0.28~ & 0.30~ \color{equal}{$\rightarrow$}  & 0.28~ & 0.26~ \color{equal}{$\rightarrow$}  & 0.28~ & 0.27~ \color{equal}{$\rightarrow$}  & 0.29~ & 0.26~ \color{bad}{$\searrow$}  & 0.27~ & 0.22~ \color{bad}{$\searrow$}  & \textbf{0.31}~ & 0.13~ \color{superbad}{$\downarrow$}  & \textbf{0.31}~ & 0.13~ \color{superbad}{$\downarrow$}  \\ 
Gardens & day\hbox{-}right & day\hbox{-}left & \textbf{0.77}~ & 0.66~ \color{bad}{$\searrow$}  & \textbf{0.77}~ & 0.74~ \color{equal}{$\rightarrow$}  & \textbf{0.77}~ & 0.67~ \color{bad}{$\searrow$}  & \textbf{0.77}~ & 0.75~ \color{equal}{$\rightarrow$}  & \textbf{0.77}~ & 0.72~ \color{equal}{$\rightarrow$}  & \textbf{0.77}~ & 0.73~ \color{equal}{$\rightarrow$}  & 0.76~ & 0.72~ \color{equal}{$\rightarrow$}  & \textbf{0.77}~ & 0.67~ \color{bad}{$\searrow$}  & 0.71~ & 0.55~ \color{bad}{$\searrow$}  & 0.71~ & 0.55~ \color{bad}{$\searrow$}  \\ 
Gardens & day\hbox{-}right & night\hbox{-}right & 0.80~ & 0.74~ \color{equal}{$\rightarrow$}  & 0.80~ & 0.79~ \color{equal}{$\rightarrow$}  & 0.80~ & 0.78~ \color{equal}{$\rightarrow$}  & 0.81~ & 0.79~ \color{equal}{$\rightarrow$}  & 0.81~ & 0.80~ \color{equal}{$\rightarrow$}  & 0.81~ & 0.79~ \color{equal}{$\rightarrow$}  & \textbf{0.82}~ & 0.79~ \color{equal}{$\rightarrow$}  & 0.80~ & 0.73~ \color{equal}{$\rightarrow$}  & 0.78~ & 0.64~ \color{bad}{$\searrow$}  & 0.78~ & 0.63~ \color{bad}{$\searrow$}  \\ 
\end{tabular}
\label{tab:pr_results}
\end{table}
\end{landscape}

\section{Summary and Conclusion}
\label{sec:conclusion}
We discussed and evaluated available VSA implementations theoretically and experimentally. 
We created a general overview of the most important properties and provided insights especially to the various implemented binding operators (taxonomy of Fig.~\ref{fig: taxonomy}). 
It was shown that self-inverse binding operations benefit in applications such as analogical reasoning ("What is the Dollar of Mexico?").
On the other hand, these self-inverse architectures, like \textbf{MAP-B} and \textbf{MAP-C}, show a trade-off between an exactly working binding (by using a binary vectors space like $\{0,1\}$ or $\{-1,1\}$) or a high bundling capacity (by using real-valued vectors). 
In the bundling capacity experiment, the sparse binary VSA \textbf{BSDC} performed well and required only a small number of dimensions. 
However, in combination with binding, the required number of dimensions increased significantly (and also including the thinning procedure did not improve this result).
Regarding the real-world application to place recognition, the sparse VSAs did not perform as well as other VSAs. 
Presumably, this can be improved by a different encoding approach or by using a higher number of dimensions (which would be feasible given the storage efficiency of sparse representations).
High performance at both synthetic and real-world experiments could be observed in the simplified complex architecture \textbf{FHRR} that uses only the angles of the complex values.
Since this architecture is not self-inverse, it requires a separate unbinding operation and cannot solve the ``What is the dollar of Mexico?" example by Kanerva's elegant approach. However, it could presumably be solved using other methods that iteratively process the knowledge tree (e.g., the readout machine in \cite{Plate1995}), but come at increased computational costs.
Furthermore, the two matrix binding VSAs (\textbf{MABT} and \textbf{VTB}) also show good results in the practical applications of language and place recognition. 
However, the drawback of these architecture is the high computational effort for binding. 

This paper, in particular the taxonomy of binding operations, revealed a very large diversity in available VSAs and the necessity of continued efforts to systematize these approaches.
However, the theoretical insights from this paper together with the provided experimental results on synthetic and real data can be used to select an appropriate VSA for new applications.
Further, they are hopefully also useful for the development of new VSAs.

Although the memory consumption and computational costs per dimension can significantly vary between VSAs, the experimental evaluation compared different VSAs using a common number of dimensions. 
We made this decision since the actual costs depend on several factors like the underlying hard- and software, or the required computational precision for the current task. For example, some high-level languages like Matlab do not well support binary representations and not all CPUs support half-precision floats. We consider the number of dimensions as an intuitive common basis for comparison between VSAs that can later be converted to memory consumption and computational costs once the influencing factors for a particular application are clear.
Recent in-memory implementations of VSA operators \cite{Karunaratne2020} are important steps towards VSA specific hardware.
Nevertheless, a more in-depth evaluation of resource consumption of the different VSAs is a very important part of future work. However, this will require additional design decisions and assumptions about properties of the underlying hard- and software.

Finally, we want to repeat the importance of permutations for VSAs. However, as explained in Sec.~\ref{sec:vsa_props}, we decided to not particularly evaluate differences in combination with permutations since they are applied very similarly in all VSAs (however, simple permutations were used in the language recognition task).


%
%

\bibliographystyle{spmpsci}      
\bibliography{library}

\begin{thebibliography}{10}
\providecommand{\url}[1]{{#1}}
\providecommand{\urlprefix}{URL }
\expandafter\ifx\csname urlstyle\endcsname\relax
  \providecommand{\doi}[1]{DOI~\discretionary{}{}{}#1}\else
  \providecommand{\doi}{DOI~\discretionary{}{}{}\begingroup
  \urlstyle{rm}\Url}\fi

\bibitem{Ahmad15}
Ahmad, S., Hawkins, J.: Properties of sparse distributed representations and
  their application to hierarchical temporal memory.
\newblock CoRR  (2015)

\bibitem{Ahmad2019}
Ahmad, S., Scheinkman, L.: {How Can We Be So Dense? The Benefits of Using
  Highly Sparse Representations}.
\newblock CoRR  (2019)

\bibitem{cmu}
{Badino}, H., {Huber}, D., {Kanade}, T.: Visual topometric localization.
\newblock In: Proc. of Intelligent Vehicles Symp. (2011)

\bibitem{Bellman61}
Bellman, R.E.: {Adaptive Control Processes: A Guided Tour}.
\newblock MIT Press (1961)

\bibitem{Beyer99}
Beyer, K., Goldstein, J., Ramakrishnan, R., Shaft, U.: When is ``nearest
  neighbor'' meaningful?
\newblock In: Database Theory --- ICDT'99, pp. 217--235. Springer Berlin
  Heidelberg, Berlin, Heidelberg (1999)

\bibitem{Cheung19}
Cheung, B., Terekhov, A., Chen, Y., Agrawal, P., Olshausen, B.: Superposition
  of many models into one.
\newblock In: Advances in Neural Information Processing Systems 32, pp.
  10868--10877. Curran Associates, Inc. (2019)

\bibitem{Danihelka16}
Danihelka, I., Wayne, G., Uria, B., Kalchbrenner, N., Graves, A.: Associative
  long short-term memory.
\newblock In: Proceedings of The 33rd International Conference on Machine
  Learning, vol.~48, pp. 1986--1994. PMLR, New York, USA (2016)

\bibitem{Eliasmith2013}
Eliasmith, C.: {How to Build a Brain: A Neural Architecture for Biological
  Cognition}.
\newblock Oxford Univ. Press (2013)

\bibitem{Frady2018}
Frady, E.P., Kleyko, D., Sommer, F.T.: {A Theory of Sequence Indexing and
  Working Memoryin Recurrent Neural Networks}.
\newblock Neural Computation \textbf{30}(6), 1449--1513 (2018).
\newblock \doi{10.1162/neco}

\bibitem{Frady2021}
Frady, E.P., Kleyko, D., Sommer, F.T.: {Variable Binding for Sparse Distributed
  Representations: Theory and Applications}.
\newblock IEEE Transactions on Neural Networks and Learning Systems pp. 1--14
  (2021).
\newblock \doi{10.1109/TNNLS.2021.3105949}.
\newblock \urlprefix\url{https://ieeexplore.ieee.org/document/9528907/}

\bibitem{Gallant2013}
Gallant, S.I., Okaywe, T.W.: Representing objects, relations, and sequences.
\newblock Neural Computation \textbf{25}, 2038--2078 (2013)

\bibitem{Gayler98}
Gayler, R.W.: Multiplicative binding, representation operators, and analogy.
\newblock In: Advances in analogy research: Integr. of theory and data from the
  cogn., comp., and neural sciences. New Bulgarian University (1998)

\bibitem{gayler03}
Gayler, R.W.: Vector symbolic architectures answer jackendoff's challenges for
  cognitive neuroscience.
\newblock In: Proc. of ICCS/ASCS Int. Conf. on Cognitive Science, pp. 133--138.
  Sydney, Australia (2003)

\bibitem{Gayler2009}
Gayler, R.W., Levy, S.D.: {A distributed basis for analogical mapping}.
\newblock New Frontiers in Analogy Research, Proceedings of the Second
  International Conference on Analogy, ANALOGY-2009 pp. 165--174 (2009)

\bibitem{gardenspoint}
Glover, A.: Day and night with lateral pose change datasets  (2014).
\newblock
  \urlprefix\url{https://wiki.qut.edu.au/display/cyphy/Day+and+Night+with+Lateral
  +Pose+Change+Datasets}

\bibitem{stlucia}
Glover, A., Maddern, W., Milford, M., Wyeth, G.: {FAB-MAP + RatSLAM:
  Appearance-based SLAM for Multiple Times of Day}.
\newblock In: Proc. of Int. Conf. on Robotics and Automation (2010)

\bibitem{Gosmann2019}
Gosmann, J., Eliasmith, C.: {Vector-Derived Transformation Binding: An Improved
  Binding Operation for Deep Symbol-Like Processing in Neural Networks}.
\newblock Neural Computation \textbf{31}, 849--869 (2019)

\bibitem{Joshi2017}
Joshi, A., Halseth, J.T., Kanerva, P.: {Language geometry using random
  indexing}.
\newblock Lecture Notes in Computer Science (including subseries Lecture Notes
  in Artificial Intelligence and Lecture Notes in Bioinformatics) \textbf{10106
  LNCS}, 265--274 (2017).
\newblock \doi{10.1007/978-3-319-52289-0_21}

\bibitem{Kanerva1996}
Kanerva, P.: {Binary Spatter-Coding of Ordered K-tuples}.
\newblock Artificial Neural Networks -- ICANN Proceedings \textbf{1112},
  869----873 (1996)

\bibitem{Kanerva09}
Kanerva, P.: Hyperdimensional computing: An introduction to computing in
  distributed representation with high-dimensional random vectors.
\newblock Cognitive Computation \textbf{1}(2), 139--159 (2009)

\bibitem{Kanerva2010}
Kanerva, P.: {What We Mean When We Say “What's the Dollar of Mexico?”:
  Prototypes and Mapping in Concept Space}.
\newblock In: AAAI Fall Symposium: Quantum Informatics for Cognitive, Social,
  and Semantic Processes, pp. 2--6 (2010)

\bibitem{Kanerva2001}
Kanerva, P., Sjoedin, G., Kristoferson, J., Karlsson, R., Levin, B., Holst, A.,
  Karlgren, J., Sahlgren, M.: {Computing with large random patterns} (2001).
\newblock
  \urlprefix\url{http://eprints.sics.se/3138/%5Cnhttp://www.rni.org/kanerva/rwi-sics.pdf}

\bibitem{Karunaratne2020}
Karunaratne, G., {Le Gallo}, M., Cherubini, G., Benini, L., Rahimi, A.,
  Sebastian, A.: {In-memory hyperdimensional computing}.
\newblock Nature Electronics \textbf{3}(6), 327--337 (2020).
\newblock \doi{10.1038/s41928-020-0410-3}

\bibitem{Karunaratne2021}
Karunaratne, G., Schmuck, M., {Le Gallo}, M., Cherubini, G., Benini, L.,
  Sebastian, A., Rahimi, A.: {Robust high-dimensional memory-augmented neural
  networks}.
\newblock Nature Communications \textbf{12}(1), 1--12 (2021).
\newblock \doi{10.1038/s41467-021-22364-0}.
\newblock \urlprefix\url{http://dx.doi.org/10.1038/s41467-021-22364-0}

\bibitem{Kelly2013}
Kelly, M.A., Blostein, D., Mewhort, D.J.: {Encoding structure in holographic
  reduced representations}.
\newblock Canadian Journal of Experimental Psychology \textbf{67}(2), 79--93
  (2013).
\newblock \doi{10.1037/a0030301}

\bibitem{Kleyko2018a}
Kleyko, D.: {Vector Symbolic Architectures and their applications}.
\newblock Ph.D. thesis, Lule{\aa} University of Technology, Lule{\aa}, Sweden
  (2018)

\bibitem{Kleyko15}
Kleyko, D., Osipov, E., Gayler, R.W., Khan, A.I., Dyer, A.G.: Imitation of
  honey bees’ concept learning processes using vector symbolic architectures.
\newblock Biologically Inspired Cognitive Architectures \textbf{14}, 57 -- 72
  (2015).
\newblock \doi{https://doi.org/10.1016/j.bica.2015.09.002}

\bibitem{Kleyko15a}
Kleyko, D., Osipov, E., Papakonstantinou, N., Vyatkin, V., Mousavi, A.: Fault
  detection in the hyperspace: Towards intelligent automation systems.
\newblock In: 2015 IEEE 13th International Conference on Industrial Informatics
  (INDIN), pp. 1219--1224 (2015).
\newblock \doi{10.1109/INDIN.2015.7281909}

\bibitem{Kleyko2020}
Kleyko, D., Rahimi, A., Gayler, R.W., Osipov, E.: {Autoscaling Bloom filter:
  controlling trade-off between true and false positives}.
\newblock Neural Computing and Applications \textbf{32}(8), 3675--3684 (2020).
\newblock \doi{10.1007/s00521-019-04397-1}.
\newblock \urlprefix\url{https://doi.org/10.1007/s00521-019-04397-1}

\bibitem{Kleyko18}
Kleyko, D., Rahimi, A., Rachkovskij, D.A., Osipov, E., Rabaey, J.M.:
  Classification and recall with binary hyperdimensional computing: Tradeoffs
  in choice of density and mapping characteristics.
\newblock IEEE Transactions on Neural Networks and Learning Systems
  \textbf{29}(12), 5880--5898 (2018).
\newblock \doi{10.1109/TNNLS.2018.2814400}

\bibitem{Kleyko2018}
Kleyko, D., Rahimi, A., Rachkovskij, D.A., Osipov, E., Rabaey, J.M.:
  {Classification and Recall With Binary Hyperdimensional Computing: Tradeoffs
  in Choice of Density and Mapping Characteristics}.
\newblock IEEE Transactions on Neural Networks and Learning Systems pp. 1--19
  (2018).
\newblock \doi{10.1109/TNNLS.2018.2814400}

\bibitem{Krizhevsky12}
Krizhevsky, A., Sutskever, I., Hinton, G.E.: Imagenet classification with deep
  convolutional neural networks.
\newblock In: Advances in Neural Information Processing Systems 25, pp.
  1097--1105. Curran Associates, Inc. (2012)

\bibitem{Laiho2015}
Laiho, M., Poikonen, J.H., Kanerva, P., Lehtonen, E.: {High-dimensional
  computing with sparse vectors}.
\newblock In: IEEE Biomedical Circuits and Systems Conference: Engineering for
  Healthy Minds and Able Bodies, BioCAS 2015 - Proceedings, pp. 1--4. IEEE
  (2015).
\newblock \doi{10.1109/BioCAS.2015.7348414}

\bibitem{robotcar}
Maddern, W., Pascoe, G., Linegar, C., Newman, P.: {1 Year, 1000km: The Oxford
  RobotCar Dataset}.
\newblock The Int. J. of Robotics Research \textbf{36}(1), 3--15 (2017).
\newblock \doi{10.1177/0278364916679498}

\bibitem{milford12}
Milford, M., Wyeth, G.F.: Seqslam: Visual route-based navigation for sunny
  summer days and stormy winter nights.
\newblock In: Proceedings of the IEEE International Conference on Robotics and
  Automation (ICRA) (2012)

\bibitem{Neubert2021b}
Neubert, P., Schubert, S.: Hyperdimensional computing as a framework for
  systematic aggregation of image descriptors.
\newblock In: Proceedings of the IEEE/CVF Conference on Computer Vision and
  Pattern Recognition, pp. 16938--16947 (2021).
\newblock \doi{10.1109/CVPR46437.2021.01666}

\bibitem{Neubert2019a}
Neubert, P., Schubert, S., Protzel, P.: {A Neurologically Inspired Sequence
  Processing Model for Mobile Robot Place Recognition}.
\newblock IEEE Robotics and Automation Letters \textbf{4}(4), 3200--3207
  (2019).
\newblock \doi{10.1109/LRA.2019.2927096}

\bibitem{Neubert2019}
Neubert, P., Schubert, S., Protzel, P.: {An Introduction to High Dimensional
  Computing for Robotics}.
\newblock German Journal of Artificial Intelligence Special Issue:
  Reintegrating Artificial Intelligence and Robotics, Springer  (2019)

\bibitem{Neubert2021a}
Neubert, P., Schubert, S., Schlegel, K., Protzel, P.: Vector semantic
  representations as descriptors for visual place recognition.
\newblock In: Proc. of Robotics: Science and Systems (RSS) (2021).
\newblock \doi{10.15607/RSS.2021.XVII.083}

\bibitem{Osipov17}
{Osipov}, E., {Kleyko}, D., {Legalov}, A.: {Associative synthesis of finite
  state automata model of a controlled object with hyperdimensional computing}.
\newblock In: IECON 2017 - 43rd Annual Conference of the IEEE Industrial
  Electronics Society, pp. 3276--3281 (2017).
\newblock \doi{10.1109/IECON.2017.8216554}

\bibitem{Plate94Phd}
Plate, T.A.: Distributed representations and nested compositional structure.
\newblock Ph.D. thesis, University of Toronto, Toronto, Ont., Canada, Canada
  (1994)

\bibitem{Plate1995}
Plate, T.A.: {Holographic Reduced Representations}.
\newblock IEEE Transactions on Neural Networks \textbf{6}(3), 623--641 (1995).
\newblock \doi{10.1109/72.377968}

\bibitem{Plate1997}
Plate, T.A.: {A common framework for distributed representation schemes for
  compositional structure}.
\newblock Connectionist Systems for Knowledge Representations and Deduction
  (July), 15--34 (1997)

\bibitem{Plate2003}
Plate, T.A.: Holographic Reduced Representation: Distributed Representation for
  Cognitive Structures.
\newblock CSLI Publications, USA (2003)

\bibitem{Rachkovskij2001}
Rachkovskij, D.A.: {Representation and processing of structures with binary
  sparse distributed codes}.
\newblock IEEE Transactions on Knowledge and Data Engineering \textbf{13}(2),
  261--276 (2001).
\newblock \doi{10.1109/69.917565}

\bibitem{Rachkovskij2001a}
Rachkovskij, D.A., Kussul, E.M.: {Binding and normalization of binary sparse
  distributed representations by context-dependent thinning}.
\newblock Neural Computation \textbf{13}(2), 411--452 (2001).
\newblock \doi{10.1162/089976601300014592}

\bibitem{Rachkovskij12}
Rachkovskij, D.A., Slipchenko, S.V.: {Similarity-based retrieval with
  structure-sensitive sparse binary distributed representations}.
\newblock Computational Intelligence \textbf{28}(1), 106--129 (2012).
\newblock \doi{10.1111/j.1467-8640.2011.00423.x}

\bibitem{Rahimi2017}
{Rahimi}, A., {Datta}, S., {Kleyko}, D., {Frady}, E.P., {Olshausen}, B.,
  {Kanerva}, P., {Rabaey}, J.M.: {High-Dimensional Computing as a Nanoscalable
  Paradigm}.
\newblock IEEE Transactions on Circuits and Systems I: Regular Papers
  \textbf{64}(9), 2508--2521 (2017).
\newblock \doi{10.1109/TCSI.2017.2705051}

\bibitem{Schubert2020}
Schubert, S., Neubert, P., Protzel, P.: {Unsupervised Learning Methods for
  Visual Place Recognition in Discretely and Continuously Changing
  Environments}.
\newblock In: Intl. Conf. on Robotics and Automation (ICRA) (2020)

\bibitem{smolensky90}
Smolensky, P.: Tensor product variable binding and the representation of
  symbolic structures in connectionist systems.
\newblock Artif. Intell. \textbf{46}(1-2), 159--216 (1990)

\bibitem{nordland}
S{\"u}nderhauf, N., Neubert, P., Protzel, P.: Are we there yet? challenging
  seqslam on a 3000 km journey across all four seasons.
\newblock Proc. of Workshop on Long-Term Autonomy at the Int. Conf. on Robotics
  and Automation  (2013)

\bibitem{Sunderhauf2015}
S{\"{u}}nderhauf, N., Shirazi, S., Dayoub, F., Upcroft, B., Milford, M.: {On
  the performance of ConvNet features for place recognition}.
\newblock IEEE International Conference on Intelligent Robots and Systems pp.
  4297--4304 (2015).
\newblock \doi{10.1109/IROS.2015.7353986}

\bibitem{Thrun05}
Thrun, S., Burgard, W., Fox, D.: {Probabilistic Robotics (Intelligent Robotics
  and Autonomous Agents)}.
\newblock The MIT Press (2005)

\bibitem{Tissera2014}
Tissera, M.D., McDonnell, M.D.: {Enabling 'question answering' in the MBAT
  vector symbolic architecture by exploiting orthogonal random matrices}.
\newblock Proceedings - 2014 IEEE International Conference on Semantic
  Computing, ICSC 2014 pp. 171--174 (2014).
\newblock \doi{10.1109/ICSC.2014.38}

\bibitem{Widdows04}
Widdows, D.: Geometry and Meaning.
\newblock Center for the Study of Language and Information Stanford, CA (2004)

\bibitem{Widdows15}
Widdows, D., Cohen, T.: Reasoning with vectors: A continuous model for fast
  robust inference.
\newblock Logic journal of the IGPL / Interest Group in Pure and Applied Logics
  (2), 141–--173 (2015)

\bibitem{Yerxa18}
Yerxa, T., Anderson, A., Weiss, E.: {The Hyperdimensional Stack Machine}.
\newblock In: Poster at Cognitive Computing (2018)

\bibitem{Yilmaz2015}
Yilmaz, O.: {Symbolic computation using cellular automata-based
  hyperdimensional computing}.
\newblock Neural Computation \textbf{27}(12), 2661--2692 (2015).
\newblock \doi{10.1162/NECO_a_00787}

\end{thebibliography}

\end{document}